\documentclass[12pt]{article}

\usepackage{geometry}
\usepackage{titlesec}
\titleformat{\section}
  {\normalfont\normalsize\bfseries\MakeUppercase}{\thesection.}{1em}{}
\titleformat{\subsection}
  {\normalfont\normalsize\bfseries}{\thesubsection.}{1em}{}

\usepackage{graphicx}
\usepackage[fleqn]{amsmath}
\usepackage{amsfonts}
\usepackage{amssymb}
\usepackage{float}
\usepackage[dvips]{epsfig}
\usepackage{epstopdf}
\usepackage{rotating}
\usepackage{graphicx}
\usepackage{microtype}
\usepackage{url}
\usepackage{subfiles}
\usepackage{caption}
\usepackage{subcaption}
\usepackage[section]{placeins}
\usepackage{xcolor}
\usepackage{framed}
\colorlet{shadecolor}{blue!20}
\usepackage{tabulary}
\usepackage[hidelinks]{hyperref}
\usepackage[capitalize,nameinlink]{cleveref}
\usepackage{flexisym}

\usepackage{amsmath,amsthm,amssymb}

\usepackage{calrsfs}
\DeclareMathAlphabet{\pazocal}{OMS}{zplm}{m}{n}

\usepackage{multirow, bigdelim}

\usepackage{tikz}
\usetikzlibrary{decorations.pathmorphing}
\usetikzlibrary{decorations.markings}
\usetikzlibrary{arrows.meta,bending}
\usetikzlibrary{arrows.meta}
\usepackage{bm}

\usepackage{adjustbox}

\newcommand\norm[1]{\left\lVert#1\right\rVert}

\newcommand\blfootnote[1]{%
  \begingroup
  \renewcommand\thefootnote{}\footnote{#1}%
  \addtocounter{footnote}{-1}%
  \endgroup
}
\graphicspath{{./Figures/}}

\begin{document}

\vspace*{5\baselineskip}

\section*{Abstract}
\blfootnote{George Tsialiamanis, Dynamics Research Group, The University of Sheffield, Sheffield} A major problem of structural health monitoring (SHM) has been the prognosis of damage and the definition of the remaining useful life of a structure. Both tasks depend on many parameters, many of which are often uncertain. Many models have been developed for the aforementioned tasks but they have been either deterministic or stochastic with the ability to take into account only a restricted amount of past states of the structure. In the current work, a generative model is proposed in order to make predictions about the damage evolution of structures. The model is able to perform in a population-based SHM (PBSHM) framework, to take into account many past states of the damaged structure, to incorporate uncertainties in the modelling process and to generate potential damage evolution outcomes according to data acquired from a structure. The algorithm is tested on a simulated damage evolution example and the results reveal that it is able to provide quite confident predictions about the remaining useful life of structures within a population.

\graphicspath{{./Figures/}}

\section{Introduction}
\label{sec:introduction}

Structures are a major part of everyday life and activities and therefore their performance and safety should be monitored. Structural health monitoring (SHM) is the discipline of structural dynamics that aims in monitoring structures and maintaining their condition \cite{farrar2012structural}. SHM is performed in various ways which can be summarised by Rytter's hieararchy \cite{rytter1993vibrational}:
\begin{enumerate}
    \item Is there damage (\textit{existence})?
    \item Where is the damage in the system (\textit{location})?
    \item What kind of damage is present (\textit{type/classification})?
    \item How severe is the damage (\textit{extent/severity})?
    \item How much useful (safe) life remains (\textit{prognosis})?
\end{enumerate}

The first step of Rytter's hierarchy has been widely addressed in many ways \cite{Worden1997} and their application has been quite successful. The steps of localisation and classification have also been addressed in a similar data-driven manner \cite{manson2003experimental}. However, the final two steps of the hierarchy are arguably the most difficult to deal with. In order to infer the extent of damage from measurements and estimate the remaining useful life of a structure, a thorough understanding of the existing damage mechanism is required as well as sufficiently accurate modelling of the environmental and loading conditions of the structure in the future \cite{farrar2007damage}. The task is quite complicated for both physics-based and data-driven approaches. Especially, because of the uncertainty in the future environmental and loading conditions, the model, most probably, has to be stochastic. Approaches have been developed to perform such estimations following a particle filtering approach \cite{corbetta2018optimization}. However, such approaches are based on a Markov chain assumption, i.e. the next state of the structure depends only on the current state (or a predefined number of previous states). This is a restriction, as the structure and the way that damage evolves might depend on states a few steps behind the previous measurements and the actual patent that affects the next step might even vary according to the current conditions.

In the current work, to deal with such issues, a data-driven approach is followed, which aims in creating a generative model that acts as a \textit{Turing mirror} of a structure \cite{worden2020digital} with presence of evolving damage. A model is considered a Turing mirror if it can pass the Turing test \cite{turing2009computing}. The test has two participants, the \textit{interrogator} and the \textit{oracle}. The interrogator presents questions to the oracle and the latter answers. If the interrogator is not able to distinguish whether the oracle is a human or a machine, then the oracle is considered to pass the test \cite{machida2015}. The proposed algorithm is able to operate within a \textit{population-based} SHM (PBSHM) framework. The algorithm to be used is a variation of the \textit{generative adversarial networks} (GANs) \cite{goodfellow2014generative}. The algorithm variation, which is designed to act on time-series of data is the \textit{Time-series generative adversarial networks} (TimeGANs) \cite{yoon2019time} and is designed to generate artificial time-series that look real. It is also an attempt to create sequential models that are truly generative, since autoregressive approaches so far are useful in the context of forecasting but are not generative, they are deterministic \cite{yoon2019time}. The algorithm presented herein aims at providing the users with potential outcome scenarios to the evolution of damage in a structure.
\newline 

\section{Time-series generative adversarial networks (TimeGAN)}
\label{sec:timeGAN}

Generative adversarial networks \cite{goodfellow2014generative}, where initially developed in order to generate images that look real to the human eye. The algorithm is based on the idea of adversarial training, which is a competition between too agents. The two agents in this case are two neural networks, the generator and the discriminator. The first is trying to transform random noise into image samples that look real and the second is trying to distinguish whether an image sample is indeed real or artificial. After training and because of the competition of the two, they both become better at their tasks. As a result, the generator should be able to fool the discriminator into classifying artificial images as real. It is expected that the human eye will also be fooled by the generated images and in many cases it is proved that the quality and the resemblance to reality of the generated images is impressive \cite{karras2017progressive, brock2018large}. The discriminator is often an auxiliary network that is not used for some purpose, other than to train the generator. 

The whole procedure can also be applied in order to generate common vector data, but has also been used in order to generate time-series data. Following the same scheme as with images and using the appropriate type of neural networks for time-series (recurrent neural networks), in \cite{mogren2016c} a GAN that generates music is presented. In \cite{esteban2017real}, another approach is presented, which generates time-series for medical data according to some conditional variables, similar to conditional GANs \cite{mirza2014conditional}. However, for the current work, the TimeGAN \cite{yoon2019time} algorithm was chosen to be used on order to generate the desired time-series. The specific algorithm, as will be shown, is able to learn temporal characteristics of the time-series, as well as to generate potential outcomes of some incomplete time-series. 

The TimeGAN algorithm is trained in order to learn two types of probability density functions. Firstly, the algorithm learns to generate timeseries $\bm{x}_{1:t}$, which look real. This yields the first condition that the algorithm is trying to satisfy, which is given by,
\begin{equation}
    \label{eq:obj_1}
    \min_{\hat{p}}D(p(\bm{C}, \bm{x}_{1:T})||\hat{p}(\bm{C}, \hat{\bm{x}}_{1:T}))
\end{equation}
where $\bm{C}$ is some input condition vector variable that partially controls the timeseries, $p(\bm{C}, \bm{x}_{1:T})$ is the real probability density function of the timeseries, $\hat{p}(\bm{C}, \hat{\bm{x}}_{1:T})$ is the probability density function of the artificial timeseries, $\hat{\bm{x}}_{1:T}$ are the artificial timeseries and $D$ is a distance metric of the probability density functions. By satisfying this condition, the timeseries, if considered as a whole, look real by some critic, in this case the discriminator, and, after completion of training, some human critic.

The second condition that the algorithm tries to satisfy is to learn temporal characteristics of the timeseries. This condition is based on the physics that define the potential next steps, given a set of existing steps of the timeseries. This condition shall prove quite useful, when one wants to generate potential outcomes of the timeseries, conditioned on some recorded values up to timestep $t$. The second condition is given by,
\begin{equation}
    \label{eq:obj_2}
    \min_{\hat{p}}D(p(\bm{x}_{t}|\bm{C}_{t}, \bm{x}_{1:t-1})||\hat{p}(\bm{x}_{t}|\bm{C}_{t}, \bm{x}_{1:t-1}))
\end{equation}
where $p(\bm{x}_{t}|\bm{C}_{t}, \bm{x}_{1:t-1})$ is the probability density function of the value of the timeseries for timestep $t$, given the timeseries so far $\bm{x}_{1:t-1}$ and the value of the condition variable $\bm{C}_{t}$, $\hat{p}(\bm{x}_{t}|\bm{C}_{t}, \bm{x}_{1:t-1})$ is the probability density function of the potential next steps of the timeseries, as it is generated by the algorithm and $D$ is some appropriate distance measure of the probability density functions. 

\begin{figure}
    \centering
    \begin{tikzpicture}[scale=0.60, every node/.style={transform shape}]
            
        % Random sampling space
        \draw[line width=0.5mm, fill=red, fill opacity=0.2] (-1.8, -0.5) rectangle (1.5, 3.0);
        
        \node (A) at (0.0, 0.0) [draw, line width=0.5mm, minimum height=0.5cm, minimum width=1cm] {\tiny $\bm{z} \sim \pazocal{U}(-1, 1)$};
        
        \node (B) at (0.0, 1.0) [draw, line width=0.5mm, minimum height=0.5cm, minimum width=1cm] {\tiny $\bm{z} \sim \pazocal{U}(-1, 1)$};
        
        \node (C) at (0.0, 2.5) [draw, line width=0.5mm, minimum height=0.5cm, minimum width=1cm] {\tiny $\bm{z} \sim \pazocal{U}(-1, 1)$};
        
        \draw[-{Latex[width=1mm, length=2mm]}, line width=0.5mm] (A) to (B);
        
        \path (B) -- (C) node [red, font=\LARGE, midway, sloped] {$\dots$};
        
        \node (A1) at (-1.3, 0.0) {\tiny $1$};
        \node (B1) at (-1.3, 1.0) {\tiny $2$};
        \node (C1) at (-1.3, 2.5) {\tiny $T-1$};
        
        \node (T1) at (0.0, -1.0) {Sampling space $Z_{t}$};
        
        % Real space
        \draw[line width=0.5mm, fill=blue, fill opacity=0.2] (-6.8, -0.5) rectangle (-3.5, 3.0);
        
        \node (A2) at (-5.0, 0.0) [draw, line width=0.5mm, minimum height=0.5cm, minimum width=1cm] {\tiny $\bm{x}_{1}$};
        
        \node (B2) at (-5.0, 1.0) [draw, line width=0.5mm, minimum height=0.5cm, minimum width=1cm] {\tiny $\bm{x}_{2}$};
        
        \node (C2) at (-5.0, 2.5) [draw, line width=0.5mm, minimum height=0.5cm, minimum width=1cm] {\tiny $\bm{x}_{T-1}$};
        
        \draw[-{Latex[width=1mm, length=2mm]}, line width=0.5mm] (A2) to (B2);
        
        \path (B2) -- (C2) node [red, font=\LARGE, midway, sloped] {$\dots$};
        
        \node (A12) at (-6.3, 0.0) {\tiny $1$};
        \node (B12) at (-6.3, 1.0) {\tiny $2$};
        \node (C12) at (-6.3, 2.5) {\tiny $T-1$};
        
        \node (T2) at (-5.0, -1.0) {Real samples $X_{t}$};
        
        % Embedding space
        \draw[line width=0.5mm, fill=green, fill opacity=0.2] (-4.3, 4.5) rectangle (-1.0, 8.0);
        
        \node (A3) at (-2.5, 5.0) [draw, line width=0.5mm, minimum height=0.5cm, minimum width=1cm] {\tiny $\bm{h}_{1}$};
        
        \node (B3) at (-2.5, 6.0) [draw, line width=0.5mm, minimum height=0.5cm, minimum width=1cm] {\tiny $\bm{h}_{2}$};
        
        \node (C3) at (-2.5, 7.5) [draw, line width=0.5mm, minimum height=0.5cm, minimum width=1cm] {\tiny $\bm{h}_{T-1}$};
        
        \draw[-{Latex[width=1mm, length=2mm]}, line width=0.5mm] (A3) to (B3);
        
        \path (B3) -- (C3) node [red, font=\LARGE, midway, sloped] {$\dots$};
        
        \node (A13) at (-3.8, 5.0) {\tiny $1$};
        \node (B13) at (-3.8, 6.0) {\tiny $2$};
        \node (C13) at (-3.8, 7.5) {\tiny $T-1$};
        
        \node (T3) at (-6.3, 6.25) {Embedding space $\pazocal{H}$};
        
        % \draw[-{Latex[width=2mm, length=3mm]}, line width=0.5mm] (-5.15, 3.0) to (-3.15, 4.5);
        
        \draw[-{Latex[width=2mm, length=3mm]}, line width=0.5mm] (-5.15, 3.0) -- node[above, sloped] {\scriptsize Embedding} ++(2.0, 1.5);
        
        \draw[-{Latex[width=2mm, length=3mm]}, line width=0.5mm] (-0.15, 3.0) -- node[above, sloped] {\scriptsize Generate} ++(-2.0, 1.5);
        
        % Reconstruction real space
        \draw[line width=0.5mm, fill=blue, fill opacity=0.2] (-6.8, 9.5) rectangle (-3.5, 13.0);
        
        \node (A4) at (-5.0, 10.0) [draw, line width=0.5mm, minimum height=0.5cm, minimum width=1cm] {\tiny $\bm{x}_{1}$};
        
        \node (B4) at (-5.0, 11.0) [draw, line width=0.5mm, minimum height=0.5cm, minimum width=1cm] {\tiny $\bm{x}_{2}$};
        
        \node (C4) at (-5.0, 12.5) [draw, line width=0.5mm, minimum height=0.5cm, minimum width=1cm] {\tiny $\bm{x}_{T-1}$};
        
        \draw[-{Latex[width=1mm, length=2mm]}, line width=0.5mm] (A2) to (B2);
        
        \path (B4) -- (C4) node [red, font=\LARGE, midway, sloped] {$\dots$};
        
        \node (A14) at (-6.3, 10.0) {\tiny $1$};
        \node (B14) at (-6.3, 11.0) {\tiny $2$};
        \node (C14) at (-6.3, 12.5) {\tiny $T-1$};
        
        \node (T4) at (-5.0, 13.5) {Real space $\pazocal{X}$};
        
        \draw[-{Latex[width=2mm, length=3mm]}, line width=0.5mm] (-3.15, 8.0) -- node[above, sloped] {\scriptsize Recover} ++(-2.0, 1.5);
        
        % Discriminate space
        \draw[line width=0.5mm, fill=orange, fill opacity=0.2] (-1.8, 9.5) rectangle (1.5, 10.5);
        
        \node (T5) at (-0.15, 11.0) {Discriminator};
        \node (T5) at (-0.15, 10.0) {$[0, 1]$};
        
        \draw[-{Latex[width=2mm, length=3mm]}, line width=0.5mm] (-2.15, 8.0) -- node[above, sloped] {\scriptsize Discriminate} ++(2.0, 1.5);
        
    \end{tikzpicture}
    \caption{ \centering General framework followed in order to train the TimeGAN algorithm.}
    \label{fig:timeGAN_layout}
\end{figure}
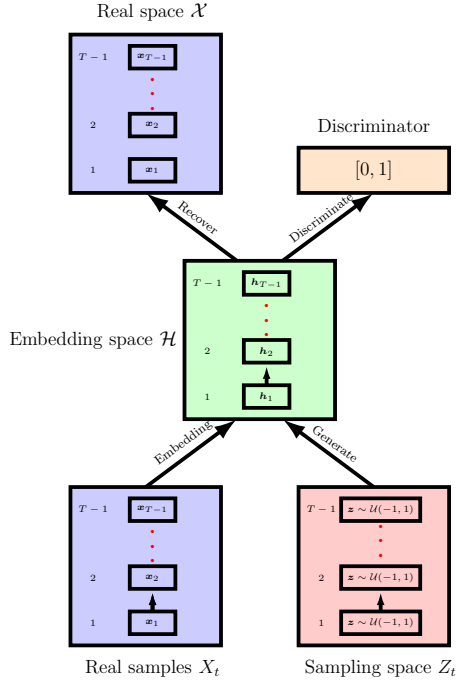

The framework followed in order to achieve a generator model that satisfies the above conditions is slightly different to the one followed in classic GANs. The main difference is that according to the TimeGAN framework, both a latent and noise space are used. The general framework is schematically shown in Figure \ref{fig:timeGAN_layout}. At first, timeseries from the real space $\bm{x}_{1:t}$ are converted into corresponding timeseries in the embedding space $\bm{h}_{1:t} \in \pazocal{H}$ using an embedding function $e$ given by,
\begin{equation}
    \bm{h}_{t} = e(\bm{C}_{t}, \bm{h}_{1:t-1}, \bm{x}_{1:t})
    \label{eq:embedding}
\end{equation}
where $\bm{C}_{t}$ are the input variables that partially control the timeseries. As described by the above equation, the embedding depends on the embeddings up to the previous timestep and on the values of all the timeseries steps so far. This forces the algorithm to learn the embeddings of the timeseries according to patterns that can even be as long as the timeseries. Together with the embedding function, a function that restores the embedded timeseries back to the real space $\pazocal{X}$. The restoration function is given by,
\begin{equation}
    \hat{\bm{x}}_{1:t} = r(\bm{h}_{1:t})
\end{equation}
where $\hat{\bm{x}}_{1:t}$ is the approximation of the reconstruction of the embedded timeseries $\bm{h}_{1:t}$. In practical applications, both functions are recurrent neural networks \cite{medsker2001recurrent} and more specifically recurrent neural networks with \textit{long-short term memory} (LSTM) units \cite{hochreiter1997long} or \textit{gated recurrent units} (GRU) \cite{chung2014empirical}. The specific types of neural networks can efficiently learn dependencies between distant timesteps of the timeseries.

In a similar manner, the generator and the discriminator of the algorithm are defined. The generator transforms timeseries from the noise space $\pazocal{Z}$ to timeseries in the embedding space $\pazocal{H}$. The equation of the generator is given by,
\begin{equation}
    \hat{\bm{h}}_{t} = g(\bm{C}_{t}, \hat{\bm{h}}_{1:t-1}, \bm{z}_{1:t})
    \label{eq:generator_equation}
\end{equation}
where $t$ is the current timestep, $g$ is the generator network, $\hat{\bm{h}}_{1:t-1}$ is the generated timeseries in the embedding space up to timestep $t-1$ and $\bm{z}_{t}$ is the noise vector which is sampled from a pre-defined distribution (in the current work, the uniform distribution is used, i.e.\ $\bm{z}_{t} \sim \pazocal{U}(-1, 1)$).

Similarly, the discriminator tries to identify whether a timeseries is real or artificial. This is done by taking as input the whole embedded timeseries $\bm{h}_{1:t}$ and the potential input variables $\bm{C}_{1:t}$. The equation of the discriminator is given by,
\begin{equation}
    y = d(\bm{C}_{1:t}, \bm{h}_{1:t})
    \label{eq:discriminator}
\end{equation}
where $d$ is the discriminator network. The prediction $y$ of the discriminator is one for timeseries that it considers to be real and zero for fake or artificial timeseries. The discriminator and the generator are also recurrent neural networks with LSTM or GRU units.  

For the purposes of satisfying the two conditions of equations (\ref{eq:obj_1}) and (\ref{eq:obj_2}), three different loss functions are used during training. The first one is defined for the training of the embedding and reconstruction of the timeseries. The reconstruction loss $\pazocal{L}_{R}$ is given by,
\begin{equation}
    \pazocal{L}_{R} = \mathbb{E}[\sum_{t}\norm{\bm{x_{t}} - \hat{\bm{x}}_{t}}_{2}]
    \label{eq:reconstruction_loss}
\end{equation}
which is essentially the mean square error between the original and the reconstructed values of the timeseries. This loss function is similar to when training an autoencoder \cite{Kramer1991}.

The second loss function is for the adversarial training part of the algorithm. It is considered an unsupervised training and its goal is to force the discriminator and the generator to perform their tasks. The unsupervised loss function $\pazocal{L}_{U}$ is given by,
\begin{equation}
    \pazocal{L}_{U} = \mathbb{E}_{\bm{x}_{1:T} \sim p}[\sum_{n} \log y_{n}] + \mathbb{E}_{\bm{x}_{1:T} \sim \hat{p}}[\sum_{n} \log(1 - \hat{y}_{n})]
    \label{eq:unsupervised_loss}
\end{equation}
where $p$ is the probability density function of the real timeseries and $\hat{p}$ is the probability denity function of the generated timeseries, which should after training match the one of the original samples.

The third loss function used is targeted to imposing the condition of equation (\ref{eq:obj_2}). The algorithm should be able to provide a range of potential next steps, given a set of observations up to some timestep. In order to impose this ability to the algorithm, the supervised loss function $\pazocal{L}_{S}$ is used and is given by,
\begin{equation}
    \pazocal{L}_{S} = \mathbb{E}_{\bm{C}, \bm{x}_{1:t} \sim p} [\sum_{t} \norm{\bm{h}_{t} - g(\bm{C}, \bm{h_{1:t-1}, \bm{z}_{t}})}_{2}]
    \label{eq:supervised_loss}
\end{equation}
where $\bm{h}_{t}$ is the embedding of the values of the timeseries at timestep $t$. This latter loss function enforces the understanding of the physics of the timeseries to the algorithm. Generated steps are random because of the random input variables $\bm{z}_{t}$ to the generator, but are also informed by the underlying physics of the timeseries up to step $t-1$, since the generator is trained using the supervised loss function $\pazocal{L}_{S}$.

For the purposes of the current work, a fourth loss function is used. This last loss function does not affect the training procedure of the algorithm, but is used to train a model $k$ that embeds real timeseries $\bm{x}_{1:t}$ into the noise space $\pazocal{Z}$. The embedding loss function $\pazocal{L}_{e}$ is given by,
\begin{equation}
    \pazocal{L}_{e} = \mathbb{E}\norm{\bm{x}_{1:t} - r(g(k(\bm{z}_{1:t})))}^{2}
    \label{eq:embedding_loss}
\end{equation}
where $r$ is the recovery model, $g$ is the generator and $k$ is the embedding model. During training using this loss function, only the model $k$ is trained, which is also a recurrent neural network. The parameters of every other model are considered constant for this later loss function. The model $k$ will be particularly useful in order to embed timeseries in the noise space and generate potential next steps for them.

By using the specific algorithm on data that reflect the evolution of damage, the models are expected to learn the physics of the damage progress. At the same time, they are expected to incorporate the uncertainty of the process, since they are generative models and are able to generate potential outcomes. The algorithm should be able to match the probability density function of the next steps of a timeseries conditioned on the observed steps, according to the real uncertainty of the procedure, as it is reflected by the sample timeseries of the dataset, which will be used to train the TimeGAN.
\newline
\section{TimeGAN for remaining useful life estimation}
\label{sec:timeGAN_remaining_life}

Approaches have been developed in order to model damage evolution under uncertainty. In \cite{worden2008prognosis} an example is given about calculating the remaining useful life of structures with cracks in them. The approach is taking into account uncertainty, but yields quite wide intervals about the potential remaining useful life and is constrained to a specific type of damage, cracks, which should also be measurable. In \cite{corbetta2018optimization}, a first-order Markov chain assumption is made in order to define a model of the damage evolution. Although such approaches for some applications suffice a common drawback is that the first-order assumption may restrict the potential of the algorithm to locate dependencies between distant timesteps.

In order to address such issues, in the current work, the TimeGAN model is studied in order to model the evolution of damage. Such a model is considered herein to be a Turing mirror of a structure \cite{worden2020digital}. A Turing mirror is a model that behaves so similarly to the real structure, that an \textit{interrogator} cannot distinguish whether the data come from real structures or from the Turing mirror model. In the case of the TimeGAN (or when training a GAN model in general) the discriminator or some human plays the role of the interrogator. Using LSTM units for the various models that are included in the TimeGAN framework, it is expected that they will be able to learn long dependencies in the data and efficiently generate potential outcomes for damage scenarios.

The algorithm is even more appealing within a PBSHM framework \cite{PBSHMMSSP1, PBSHMMSSP2, PBSHMMSSP3, PBSHMMSSP4}. Such a framework allows exploiting data from different structures in order to perform inference about some structure that sufficient data might not be present. For damage prognosis and remaining useful life estimation, this might even be the only way to perform these tasks in a data-driven manner. Data from deployment until failure of a structure are definitely not available if the structure itself has not failed. Therefore, it is necessary to exploit data from similar structures in order to perform inference without including some physics in the inference procedure. 

Following such a PBSHM framework, the form of a dataset $D$ that can be used is given by,
\begin{equation}
    D = \{(S_{i}, \bm{x}^{i}_{1:T_{i}}) \quad i = 1, 2..., N, \; T_{i} \sim p_{T}\}
    \label{eq:acquired_dataset}
\end{equation}
where $S_{i}$ is the $i$th structure, $\bm{x}^{i}_{1:T_{i}}$ is the timeseries of some damage-sensitive feature monitored, $T_{i}$ is the total lifetime of the $i$th structure, $p_{T}$ is the probability density function of the total lifetime of all structures in the population and $N$ is the number of structures in the population.

A naive approach to defining the remaining useful lifetime of the structures within the population would be to define a trivial probability density function according to the lifetimes of the structures in the dataset $D$. By doing so, one defines a form \cite{PBSHMMSSP1}, which describes a general characteristic of the population. For a more sophisticated approach, such a dataset can be used in order to train the TimeGAN, which can be considered a Turing mirror of some damage type evolution process of the population. In order to use the model for some structure under evaluation, the sequence of data of the structure up to the current timestep $t_{c}$ is required. To find the noise sequence $\bm{z}_{1:t_{c}}$, for which the generator $g$ generates the testing timeseries $\bm{x}_{1:t_{c}}$, a recurrent neural network is trained, using as a loss function the embedding loss function of equation ($\ref{eq:embedding_loss}$). Therefore, the embedding for some testing timeseries is given by,
\begin{equation}
    \bm{z}_{1:t_{c}} = k(\bm{x}_{1:t_{c}})
\end{equation}
where $k$ is the embedding model.

Since the noise embedding up to the current timestep $t_{c}$ is available, generation of potential outcomes of the current incomplete timeseries is needed. To do so, a set of noise codes $D_{\bm{z}}$ is defined following,
\begin{equation}
    D_{\bm{z}} = \{\bm{z}^{k}_{1:T_{f}} \; | \; \bm{z}^{k}_{1:t_{c}} = k(\bm{x}_{1:t_{c}}), \bm{z}^{k}_{t_{i}} \sim \pazocal{U}(-1, 1) \; \forall \; t_{i} = t_{c} + 1, t_{c} + 2..., T_{f}\}
\end{equation}
where $t_{c}$ is the current timestep of the tested structure and $T_{f}$ is a number of timesteps large enough in order for the structure to have certainly reached its failure point and $k = 1, 2..., N_{a}$ where $N_{a}$ is the number of artificially generated samples. Subsequently, the codes are used as inputs to the generator and a dataset $D^{g}_{\bm{x}}$ with potential outcomes of the currently tested timeseries is given by,
\begin{equation}
    D^{g}_{\bm{x}} = \{\bm{x}^{k}_{1:T_{f}} = r(g(\bm{z}^{k}_{1:T_{f}}))\}
\end{equation}

In the dataset defined above, every timeseries is identical up to the current timestep $t_{c}$. Now, one can study the potential outcomes in order to see how the structure might behave in the future, i.e.\ one can ask the oracle questions, instead investigating the available data from the population. Moreover, one can define a criterion based on experience or some understanding in order to define when each potential outcome reaches the end of the lifetime of the structure. By collecting all the timesteps at which each outcome reached the end of the life of the structure, a probability density function can be defined over the remaining useful lifetime. 
\newline 

\section{Application example}
\label{sec:application}

\subsection{Description of the simulated dataset}
\label{sec:dataset_description}

\begin{figure}[h!]
    \centering
        \begin{tikzpicture}[scale=0.65, every node/.style={transform shape}]
        \draw[line width=0.5mm] (-2,1) -- (-2,-1);

        \draw[line width=0.5mm] (-2, 1) -- (-2.2, 0.8);
        \draw[line width=0.5mm] (-2, 0.8) -- (-2.2, 0.6);
        \draw[line width=0.5mm] (-2, 0.6) -- (-2.2, 0.4);
        \draw[line width=0.5mm] (-2, 0.4) -- (-2.2, 0.2);
        \draw[line width=0.5mm] (-2, 0.2) -- (-2.2, 0.0);
        \draw[line width=0.5mm] (-2, 0.0) -- (-2.2, -.2);
        \draw[line width=0.5mm] (-2, -.2) -- (-2.2, -.4);
        \draw[line width=0.5mm] (-2, -.4) -- (-2.2, -.6);
        \draw[line width=0.5mm] (-2, -.6) -- (-2.2, -.8);
        \draw[line width=0.5mm] (-2, -.8) -- (-2.2, -1.0);
    
        \node (1) at (0.0, 0.0) [draw, line width=0.5mm, minimum width=1cm, minimum height=1cm] {$\bm{m_{1}}$};
        \node (2) at (2.5, 0) [draw, line width=0.5mm, minimum width=1cm, minimum height=1cm] {$\bm{m_{2}}$};
        \node (3) at (5.0, 0.0) [draw, line width=0.5mm, minimum width=1cm, minimum height=1cm] {$\bm{m_{3}}$};
        \node (4) at (7.5, 0.0) [draw, line width=0.5mm, minimum width=1cm, minimum height=1cm] {$\bm{m_{4}}$};
        
        \draw[thick, decoration={aspect=0.65, segment length=3mm,
             amplitude=0.2cm, coil}, decorate] (-2, 0) -- (-0.5, 0);
        
        \draw[thick, decoration={aspect=0.65, segment length=3mm,
             amplitude=0.2cm, coil}, decorate] (1) --(2);
        \draw[thick, decoration={aspect=0.65, segment length=3mm,
             amplitude=0.2cm, coil}, decorate] (2) --(3);
        \draw[thick, decoration={aspect=0.65, segment length=3mm,
             amplitude=0.2cm, coil}, decorate] (3) --(4);

        \draw[-{Latex[width=3mm, length=4mm]}, line width=0.5mm] (0, 0.5) -- (0, 1.5) -- (1, 1.5);
        \node[] () at (0.5, 2.0) {$F$};
        
        \node[] () at (-1.25, 1.0) {$\bm{k_{1}}$};
        \node[] () at (1.25, 1.0) {$\bm{k_{2}}$};
        \node[] () at (3.75, 1.0) {$\bm{k_{3}}$};
        \node[] () at (6.25, 1.0) {$\bm{k_{4}}$};
        
        \draw[line width=0.5mm] (0, -0.5) -- (0, -1.0) -- (-0.5, -1.0);
        \draw[line width=0.5mm] (-0.5, -0.9) -- (-0.5, -1.1);
        \draw[line width=0.5mm]  (-0.4, -1.2) -- (-0.6, -1.2) -- (-0.6, -0.8) --                             (-0.4, -0.8);
        \draw[line width=0.5mm] (-0.6, -1.0) -- (-0.9, -1.0) -- (-0.9, -1.6);
        \draw[line width=0.5mm] (-1.1, -1.6) -- (-.7, -1.6);
        
        \draw[line width=0.5mm] (2.5, -0.5) -- (2.5, -1.0) -- (2.0, -1.0);
        \draw[line width=0.5mm] (2.0, -0.9) -- (2.0, -1.1);
        \draw[line width=0.5mm]  (2.1, -1.2) -- (1.9, -1.2) -- (1.9, -0.8) --                             (2.1, -0.8);
        \draw[line width=0.5mm] (1.9, -1.0) -- (1.6, -1.0) -- (1.6, -1.6);
        \draw[line width=0.5mm] (1.4, -1.6) -- (1.8, -1.6);
        
        \draw[line width=0.5mm] (5, -0.5) -- (5, -1.0) -- (4.5, -1.0);
        \draw[line width=0.5mm] (4.5, -0.9) -- (4.5, -1.1);
        \draw[line width=0.5mm]  (4.6, -1.2) -- (4.4, -1.2) -- (4.4, -0.8) --                             (4.6, -0.8);
        \draw[line width=0.5mm] (4.4, -1.0) -- (4.1, -1.0) -- (4.1, -1.6);
        \draw[line width=0.5mm] (3.9, -1.6) -- (4.3, -1.6);
        
        \draw[line width=0.5mm] (7.5, -0.5) -- (7.5, -1.0) -- (7.0, -1.0);
        \draw[line width=0.5mm] (7.0, -0.9) -- (7.0, -1.1);
        \draw[line width=0.5mm]  (7.1, -1.2) -- (6.9, -1.2) -- (6.9, -0.8) --                             (7.1, -0.8);
        \draw[line width=0.5mm] (6.9, -1.0) -- (6.6, -1.0) -- (6.6, -1.6);
        \draw[line width=0.5mm] (6.4, -1.6) -- (6.8, -1.6);

        \node[] at (-1.25, -1.0) {$\bm{c_{1}}$};
        \node[] at (1.25, -1.0) {$\bm{c_{2}}$};
        \node[] at (3.75, -1.0) {$\bm{c_{3}}$};
        \node[] at (6.25, -1.0) {$\bm{c_{4}}$};
    
    \end{tikzpicture} 
    \caption{\centering Mass-spring system.}
    \label{fig:mass_spring}
\end{figure}
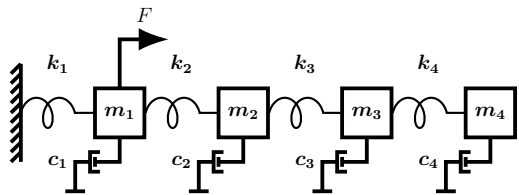

In order to evaluate how effectively can the proposed algorithm provide estimations of the remaining useful life of a structure, a simulated dataset is considered. The simulated system is shown in Figure \ref{fig:mass_spring}. In this system, damage is simulated as stiffness reductions of springs $2$ and $3$. In this case the damage sensitive feature, which is considered to be monitored, is the \textit{frequency response function} (FRF) of the acceleration of the second mass.

In order to induce some physics into the problem, damage is considered to be increasing according to some rules. A first assumption is that the structures that are studied belong to a population (a homogeneous one) of $1000$ structures. The initial and undamaged stiffness parameters of the springs are sampled from a Gaussian distribution. More specifically, $k_{2}, k_{3} \sim \mathcal{N}(6000, 120)$. For the purposes of defining the damage evolution process, a nominal step $k_{n}$ is uniformly sampled from the interval $[42, 90]$ for every structure. At every timestep of the damage evolution process, the new value of the stiffness of the second spring $k_{2}$ is defined by,
\begin{equation}
    \label{eq:damage_evol}
    k^{t+1}_{2} = k^{t}_{2} - (k_{n} + \mathcal{N}(0, 0.1 \times k_{n}))
\end{equation}
where $k^{t+1}_{2}$ is the stiffness of the next timestep and $k^{t}_{2}$ is the stiffness of the current timestep. This damage evolution process was selected with a view to having multiple sources of uncertainty that the algorithm will have to learn. The first source of uncertainty is the selection of the nominal step, which however can be approximated as the mean value of the degradation steps when several timesteps are available. The second source of uncertainty is the last term in the right hand side in equation ($\ref{eq:damage_evol}$), which cannot be learnt somehow and the algorithm will have to model it as a random variable.

\begin{figure}[!h] 
\centering
    \begin{subfigure}[b]{0.49\textwidth}
    \centering
    \includegraphics[width=\textwidth]{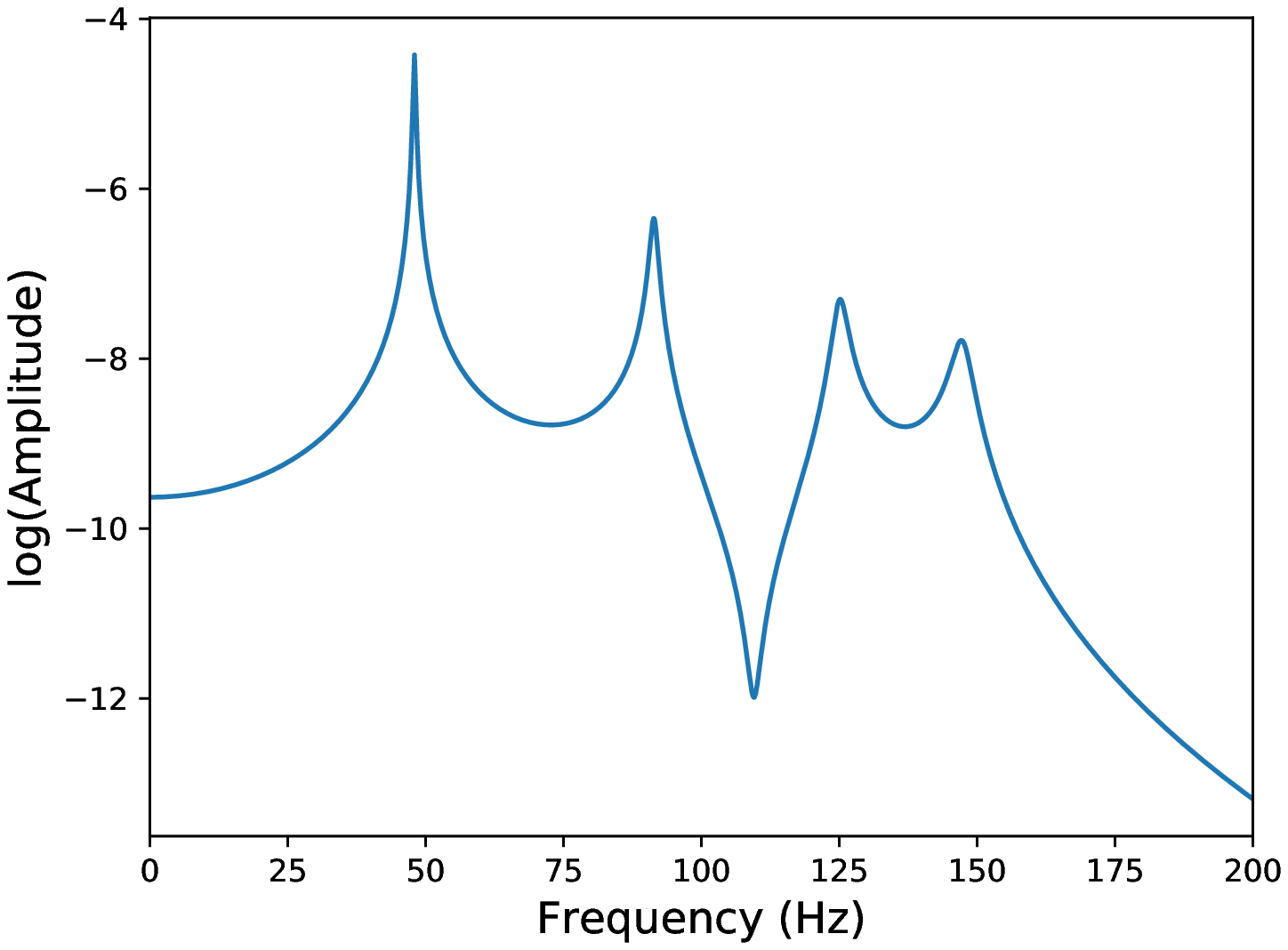}
    \caption{}
    \label{fig:orig_lines}
    \end{subfigure}
    \begin{subfigure}[b]{0.49\textwidth}
    \centering
    \includegraphics[width=\textwidth]{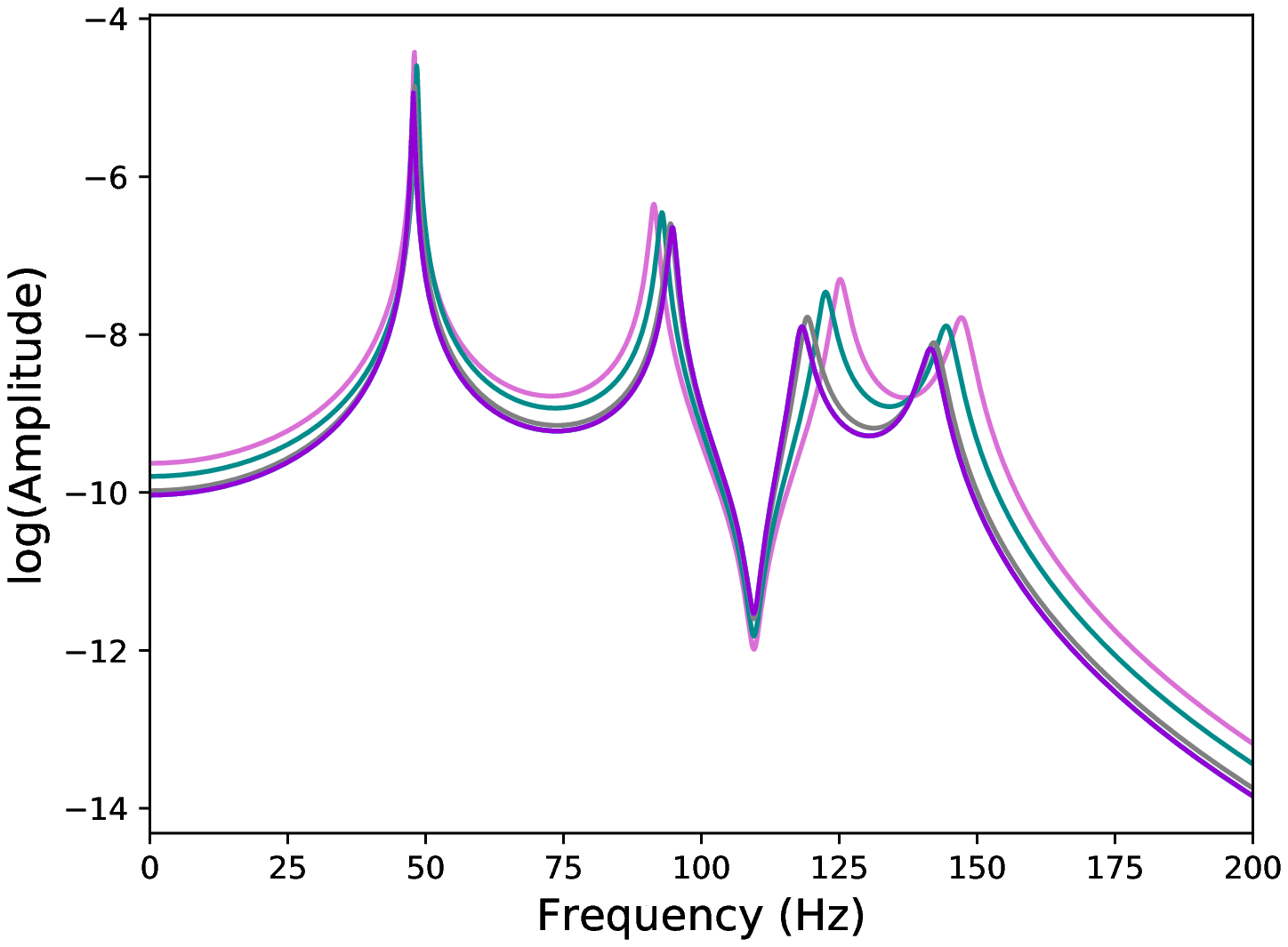}
    \caption{}
    \label{fig:target_entangled}
    \end{subfigure}
    \caption{\centering FRF sample on the beginning of the simulations (left) and FRFs with increasing damage (right), from low damage (pink) to higher stiffness reduction (purple).}
    \label{fig:sample_n_degrading_frfs}
\end{figure}

\begin{figure}[!h] 
    \centering
    \includegraphics[scale=0.45]{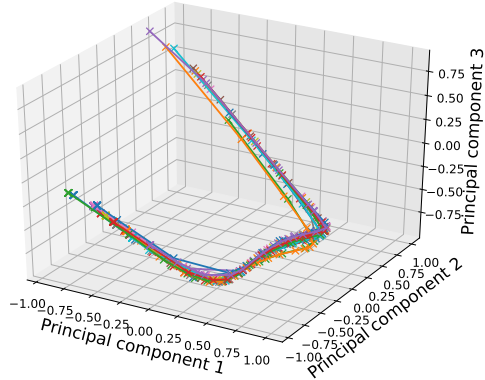}
    \caption{\centering Principal components of paths from deployment until the failure of structures within the population; different colours corespond to different structures.}
    \label{fig:path_samples_PCs}
\end{figure}

The limit state of the structures is considered to be when their third natural frequency reaches a specific point. That point here was the $119 Hz$, chosen since most structures had higher third natural frequency than that after several damage evolution timesteps (it might even be considered a conservative limit). The selected criterion is based only on human examination of the data and is aimed at simulating a criterion derived from one's experience. In Figure \ref{fig:sample_n_degrading_frfs}, FRFs of a structure as damage progresses are shown. Since the FRFs are high-dimensional features, in order to train the algorithm faster and be able to visualise the dataset, \textit{principal component analysis} (PCA) \cite{wold1987principal} was performed on the timeseries of the FRFs and the first three principal components of a subset of timeseries are shown in Figure \ref{fig:path_samples_PCs}.

\subsection{Application of the TimeGAN on the simulated dataset}
\label{sec:application_of_timeGAN}

The TimeGAN model is considered as a model to define the remaining useful life of the structures. The neural networks of the model were all chosen to be LSTM neural networks with two LSTM units. Each LSTM network had neural networks with 128 neurons in their hidden layer and all the activation functions of the networks were hyperbolic tangent activation functions, except for the activation function of the discriminator, which was a sigmoid activation function. The sampling space $\pazocal{Z}$ was selected to be a three-dimensional space and the embedding space $\pazocal{H}$ a 128-dimensional space. The real space was three-dimensional, since a PCA was performed on the FRFs before the training of the algorithm and the three first principal components explained $96\%$ of the variance of the data. In order to perform training, the Adam optimiser was used \cite{kingma2014adam}.

\begin{figure}
    \centering
    \includegraphics[scale=0.45]{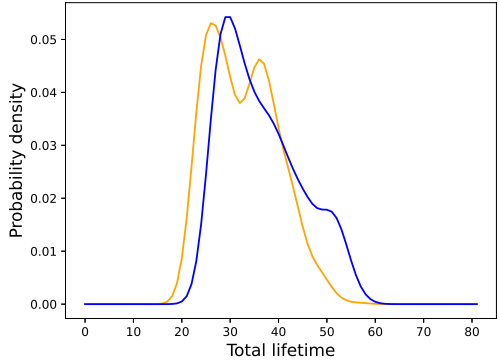}
    \caption{\centering Probability density function of the total lifetime of the real samples (blue) and of the artificial samples (orange).}
    \label{fig:synth_real_comp}
\end{figure}

A standard and overall defined validation process for training generative adversarial networks has not been established. In contrast to traditional regression or classification tasks, where a cross-validation scheme can be followed, the performance of generative adversarial networks is often evaluated according to how real do images look like or how close are the generated data distributions to the distributions of the real data. In absence of such a validation scheme, in the current work, a metric chosen to evaluate whether the TimeGAN is properly trained was the distribution of the total lifetime of the timeseries. In Figure \ref{fig:synth_real_comp}, a comparison between the PDF of the real samples and $1000$ artificially generated by the TimeGAN algorithm samples is shown. The PDF was calculated using a kernel density estimation and the bandwidth was calculated using Silverman's algorithm \cite{silverman1981using}. The KL divergence \cite{kullback1997information} between the two distributions is equal to $0.25$, which for the purposes of the current work was considered to be low enough.

\begin{figure}[] 
\centering
    \begin{subfigure}[b]{0.19\textwidth}
    \centering
    \includegraphics[width=\textwidth]{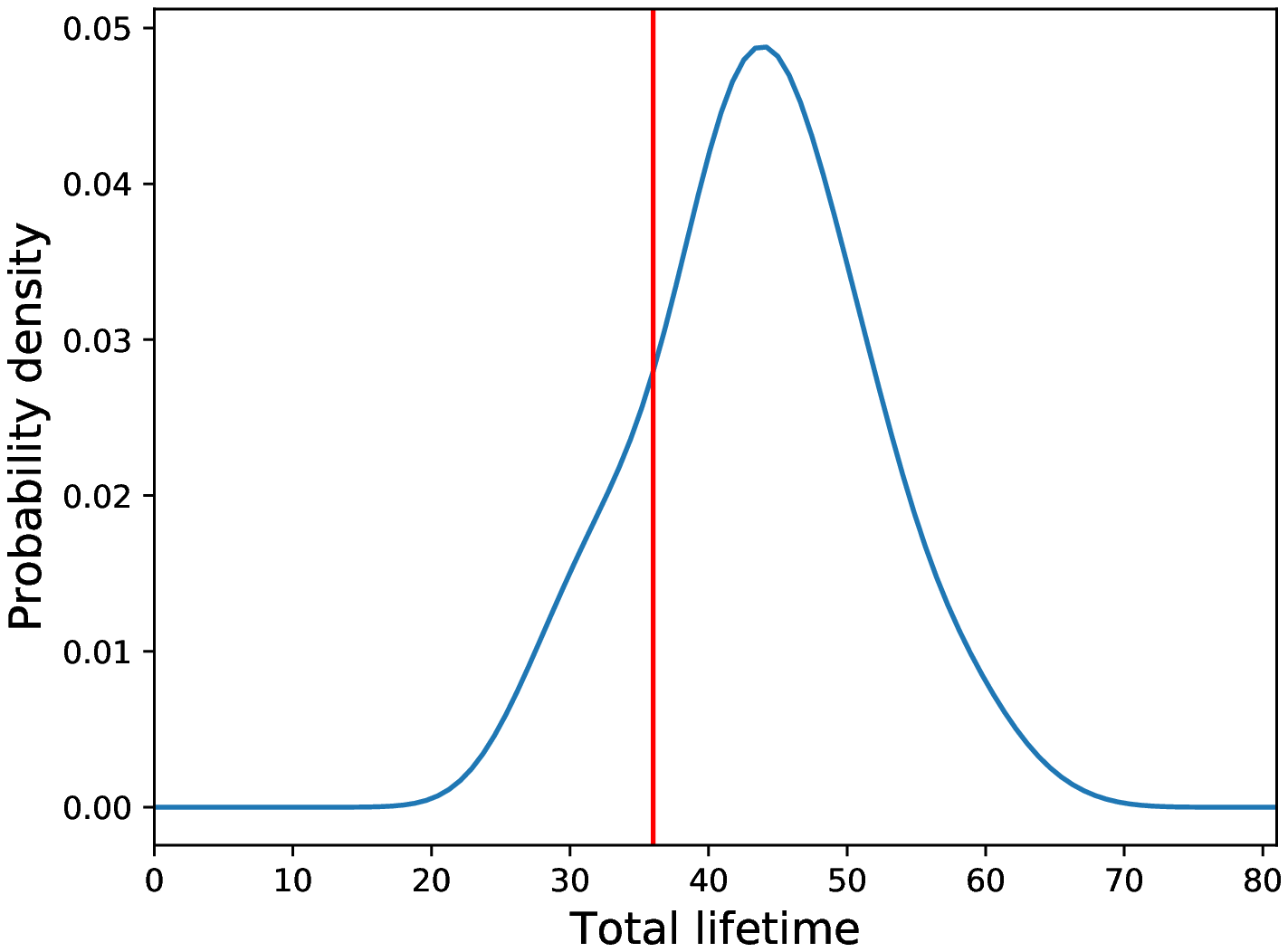}
    \caption{}
    \end{subfigure}
    \begin{subfigure}[b]{0.19\textwidth}
    \centering
    \includegraphics[width=\textwidth]{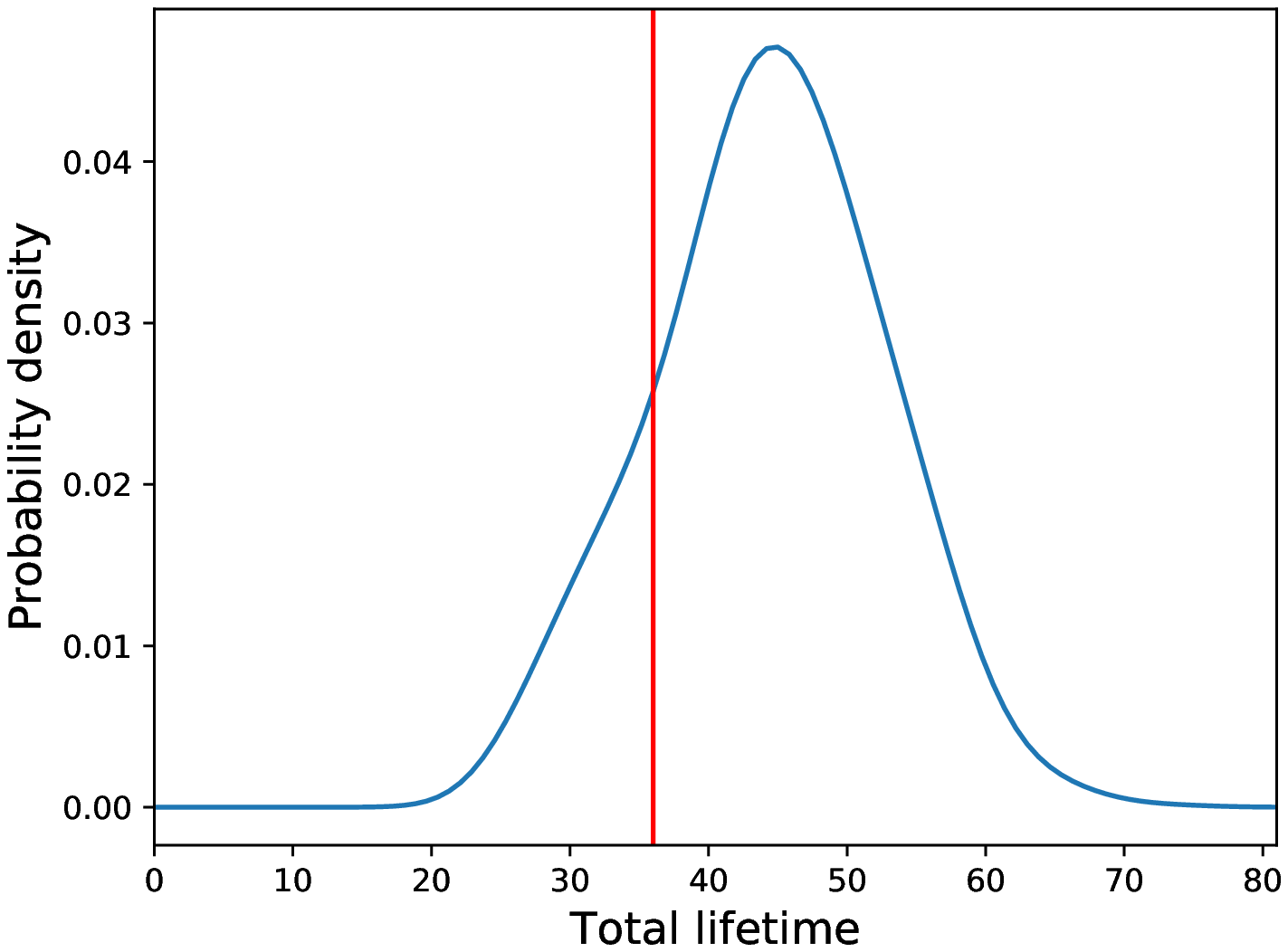}
    \caption{}
    \end{subfigure}
    \begin{subfigure}[b]{0.19\textwidth}
    \centering
    \includegraphics[width=\textwidth]{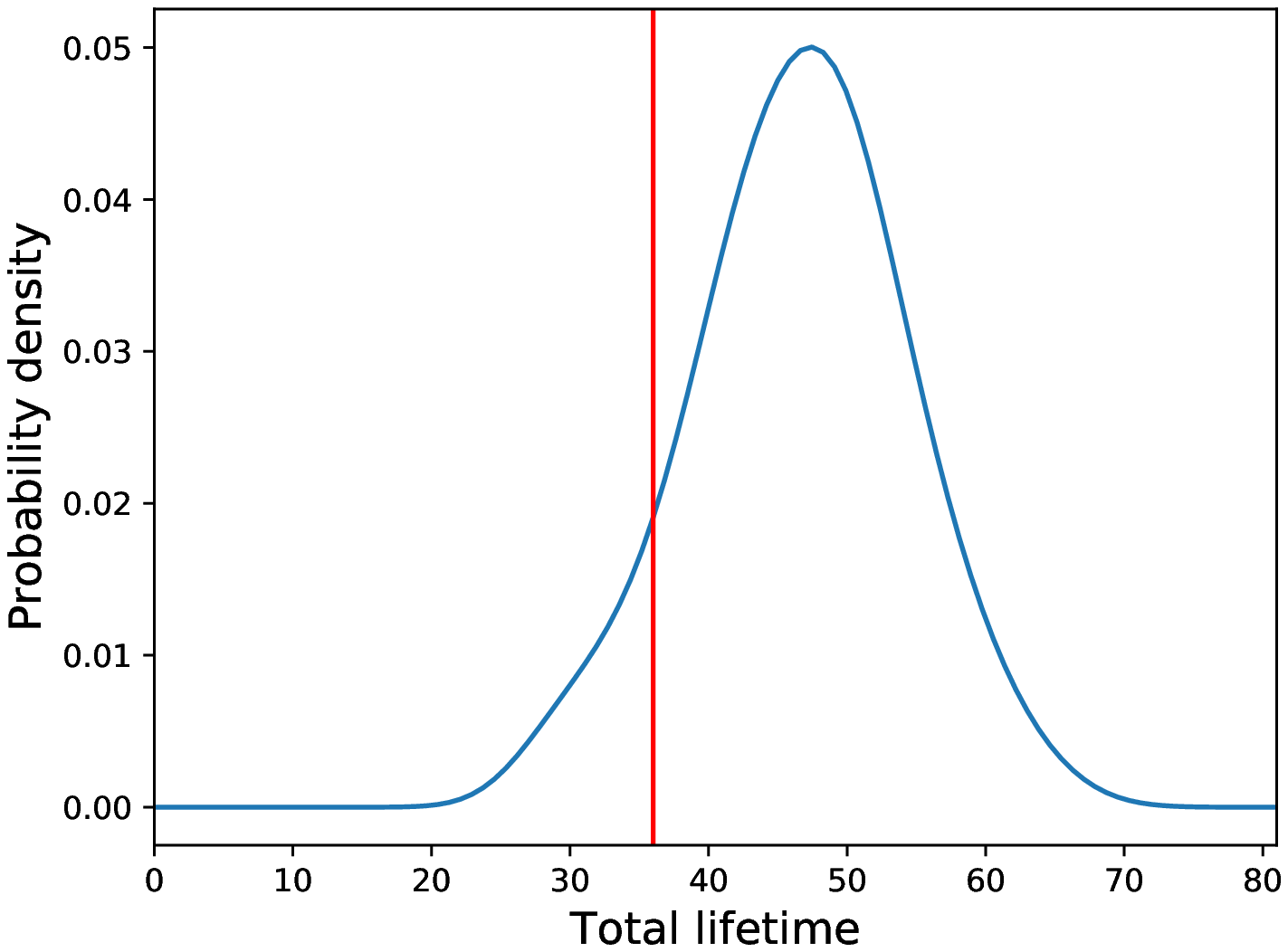}
    \caption{}
    \end{subfigure}
    \begin{subfigure}[b]{0.19\textwidth}
    \centering
    \includegraphics[width=\textwidth]{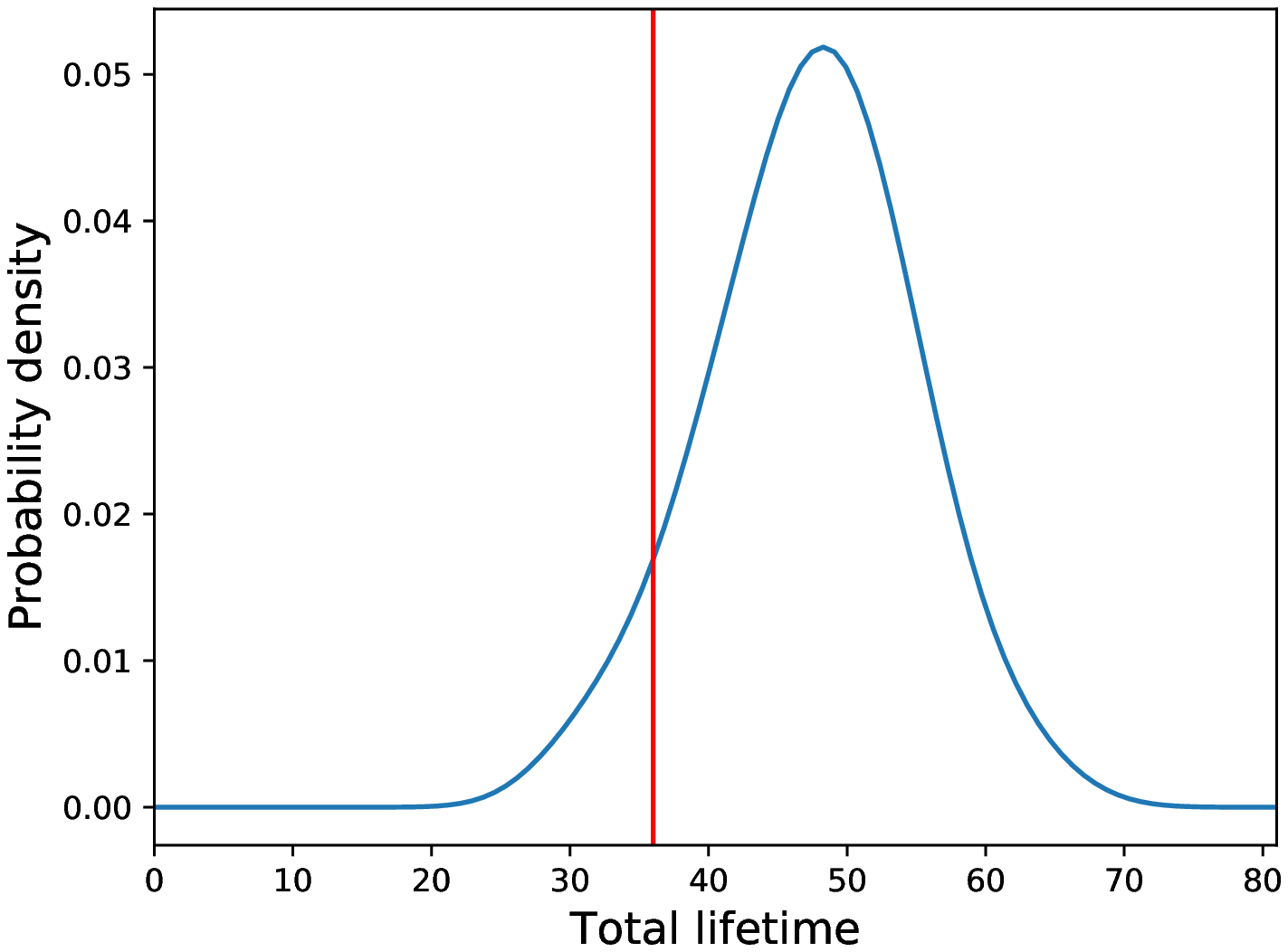}
    \caption{}
    \end{subfigure}
    \begin{subfigure}[b]{0.19\textwidth}
    \centering
    \includegraphics[width=\textwidth]{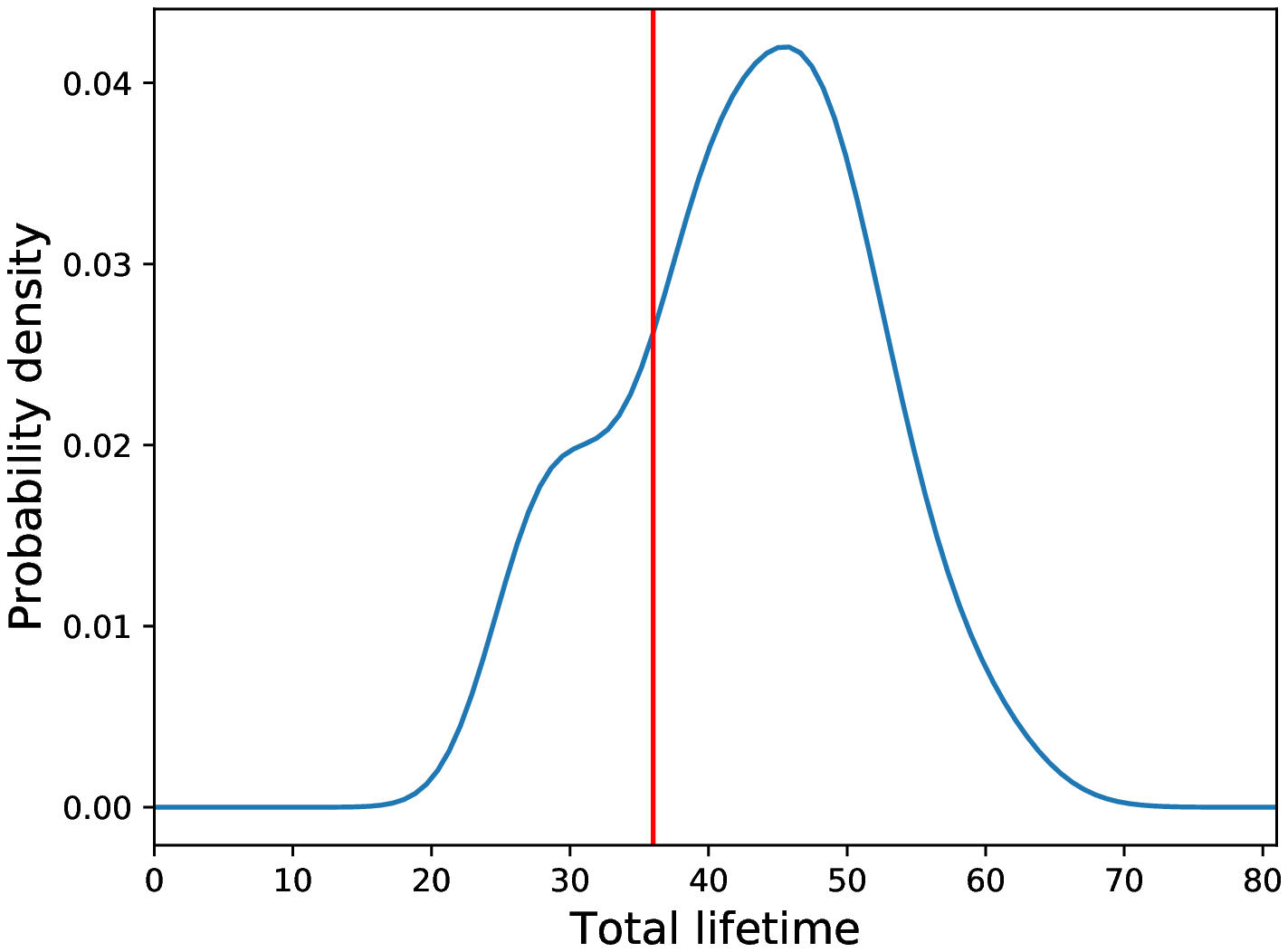}
    \caption{}
    \end{subfigure}
    \begin{subfigure}[b]{0.19\textwidth}
    \centering
    \includegraphics[width=\textwidth]{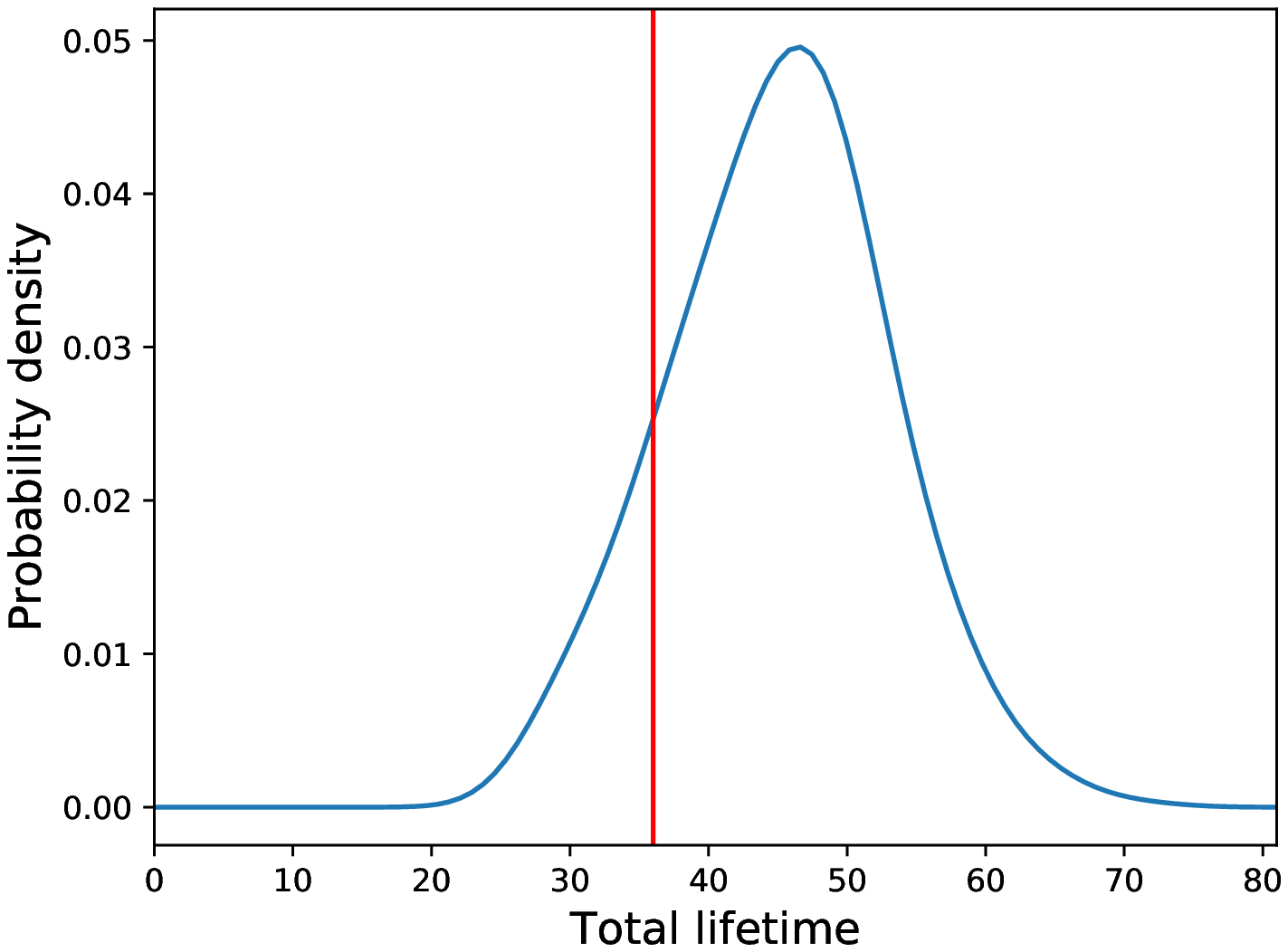}
    \caption{}
    \end{subfigure}
    \begin{subfigure}[b]{0.19\textwidth}
    \centering
    \includegraphics[width=\textwidth]{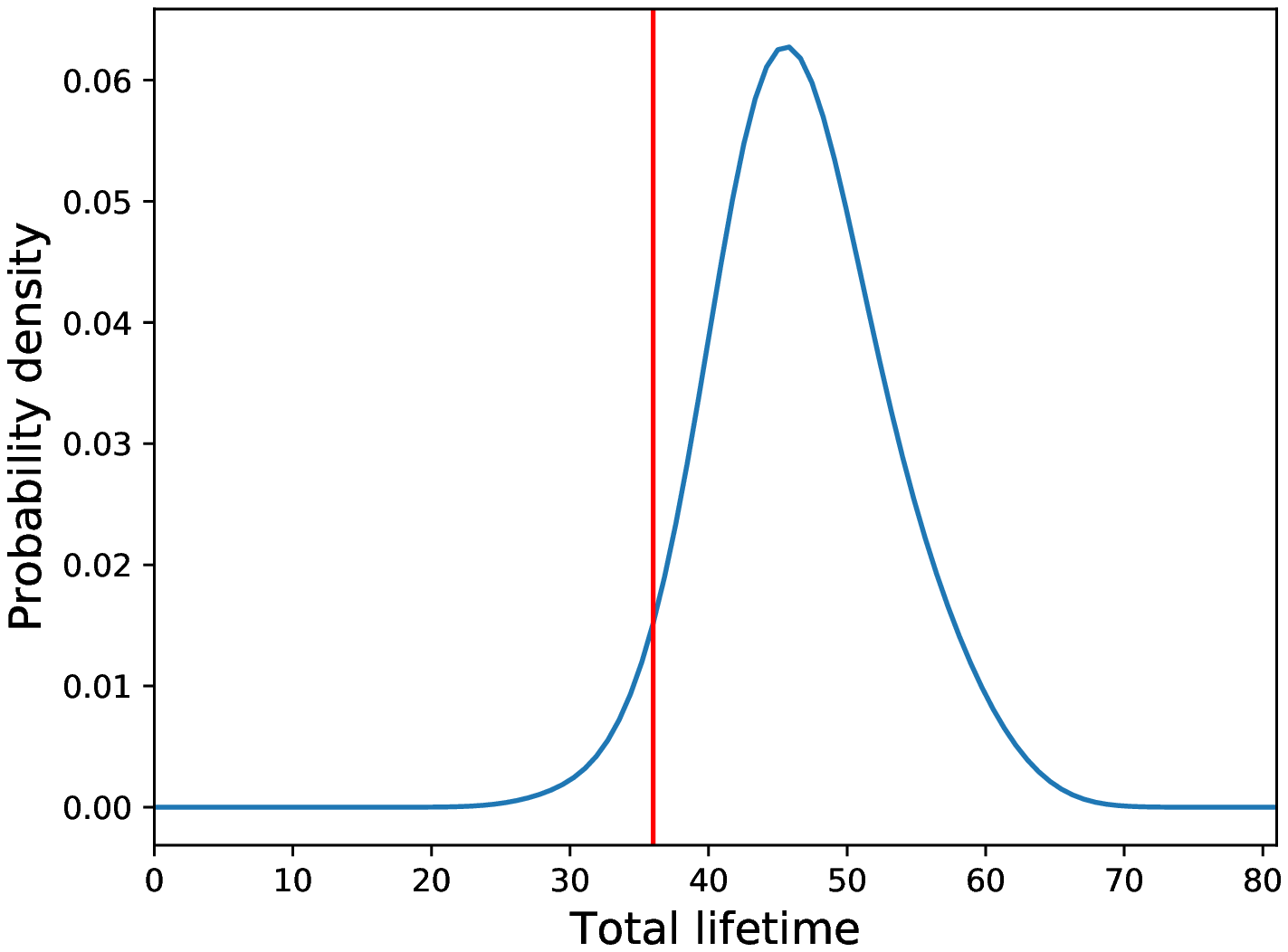}
    \caption{}
    \end{subfigure}
    \begin{subfigure}[b]{0.19\textwidth}
    \centering
    \includegraphics[width=\textwidth]{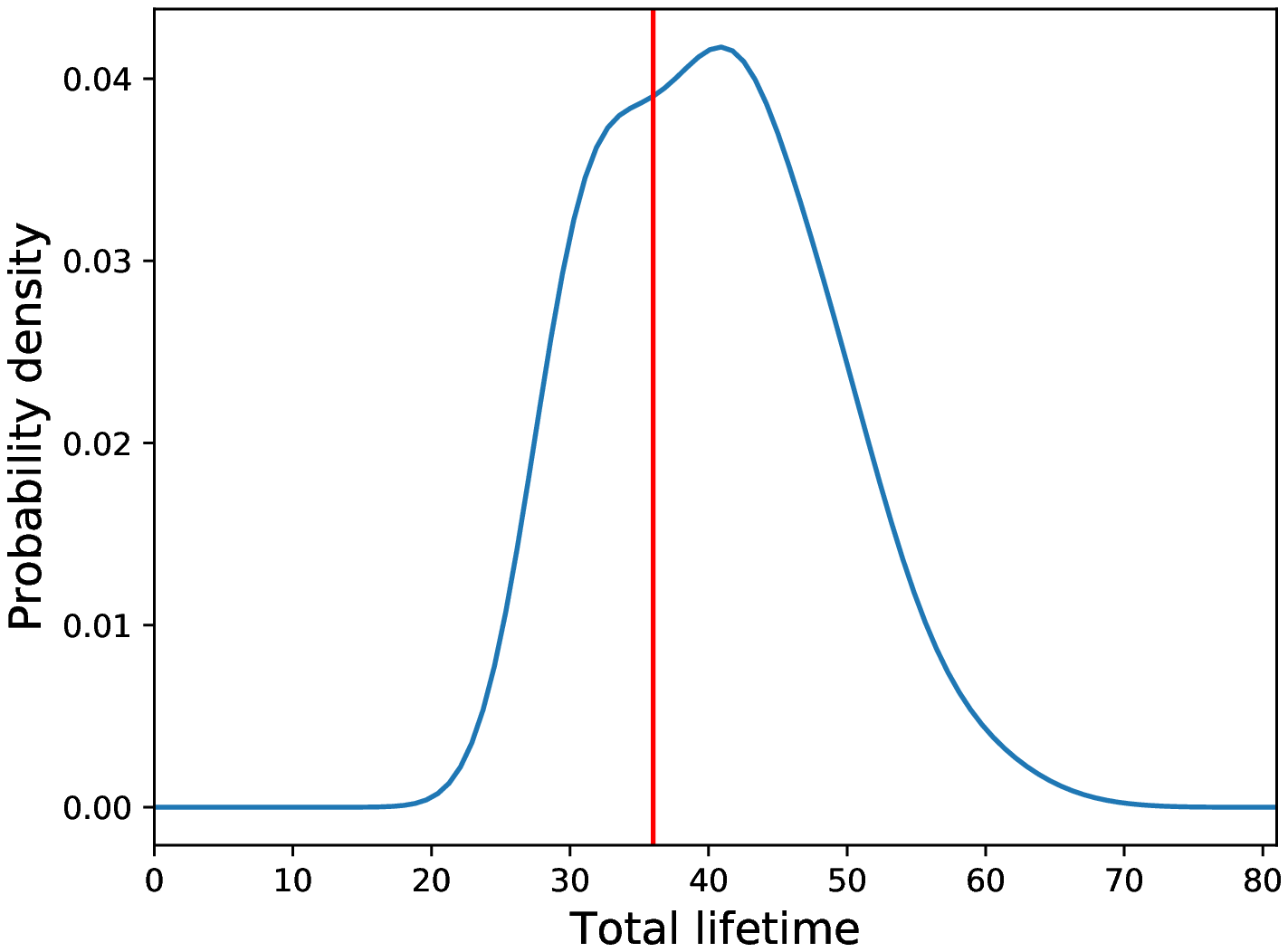}
    \caption{}
    \end{subfigure}
    \begin{subfigure}[b]{0.19\textwidth}
    \centering
    \includegraphics[width=\textwidth]{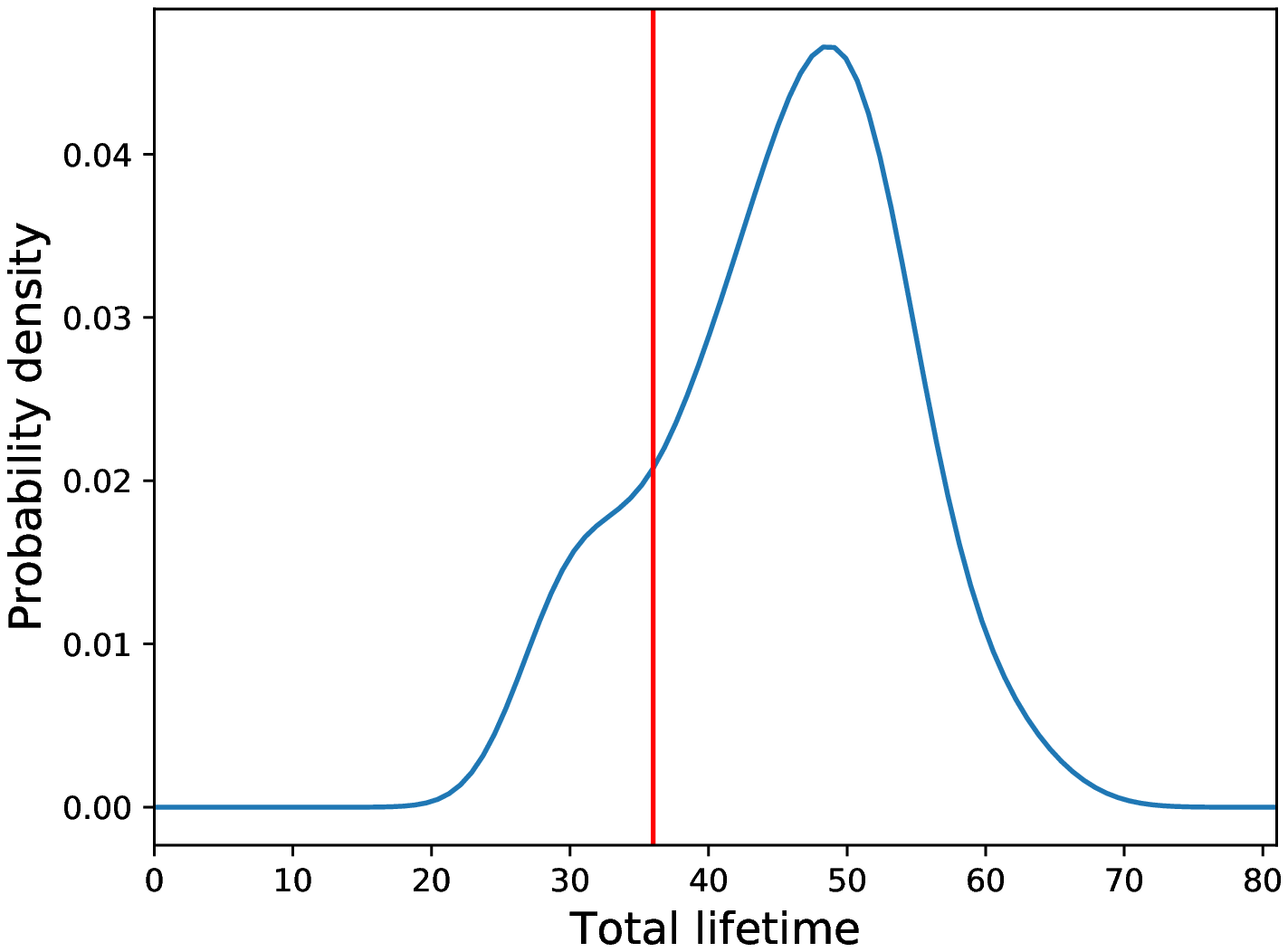}
    \caption{}
    \end{subfigure}
    \begin{subfigure}[b]{0.19\textwidth}
    \centering
    \includegraphics[width=\textwidth]{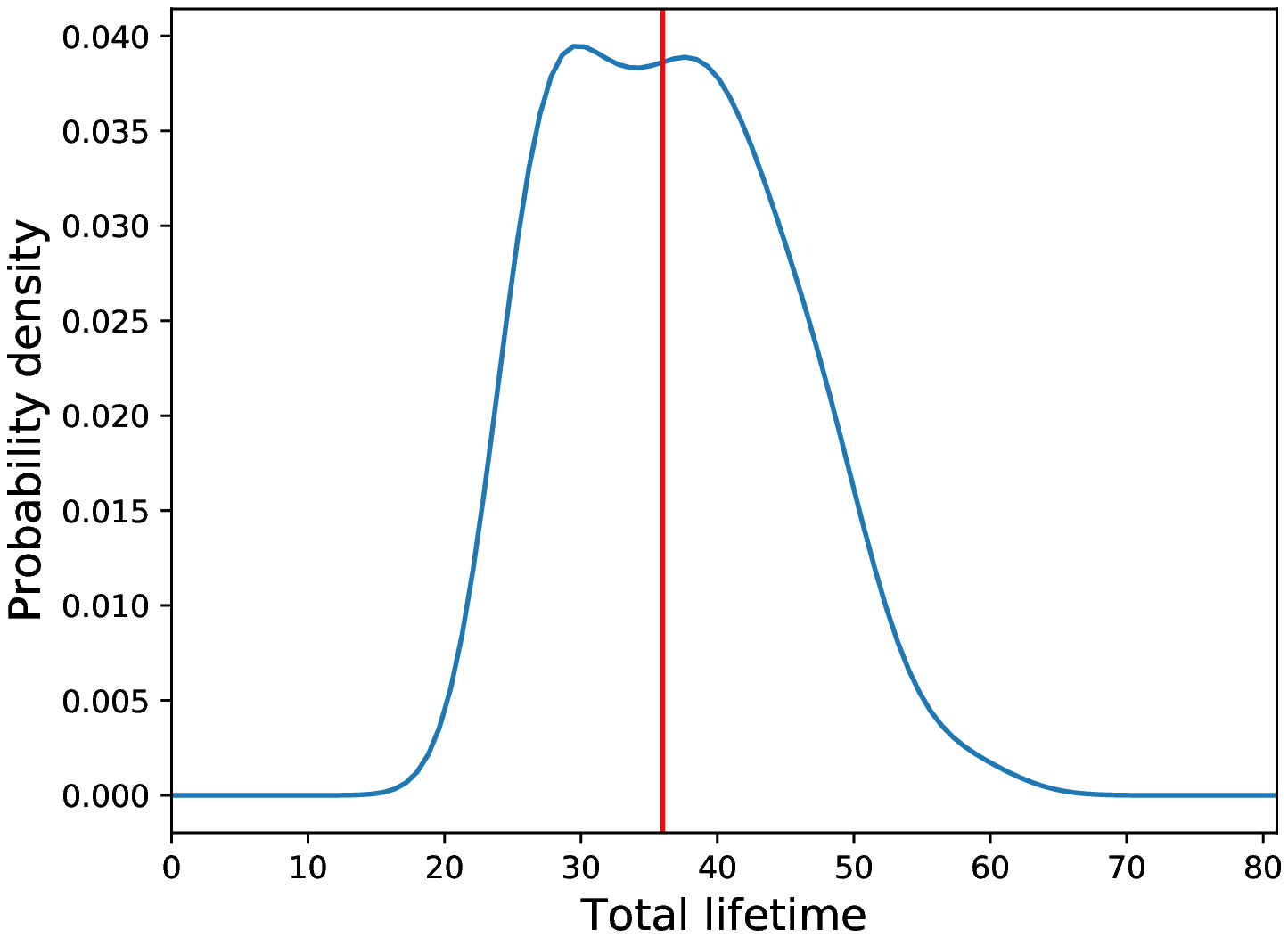}
    \caption{}
    \end{subfigure}
    \begin{subfigure}[b]{0.19\textwidth}
    \centering
    \includegraphics[width=\textwidth]{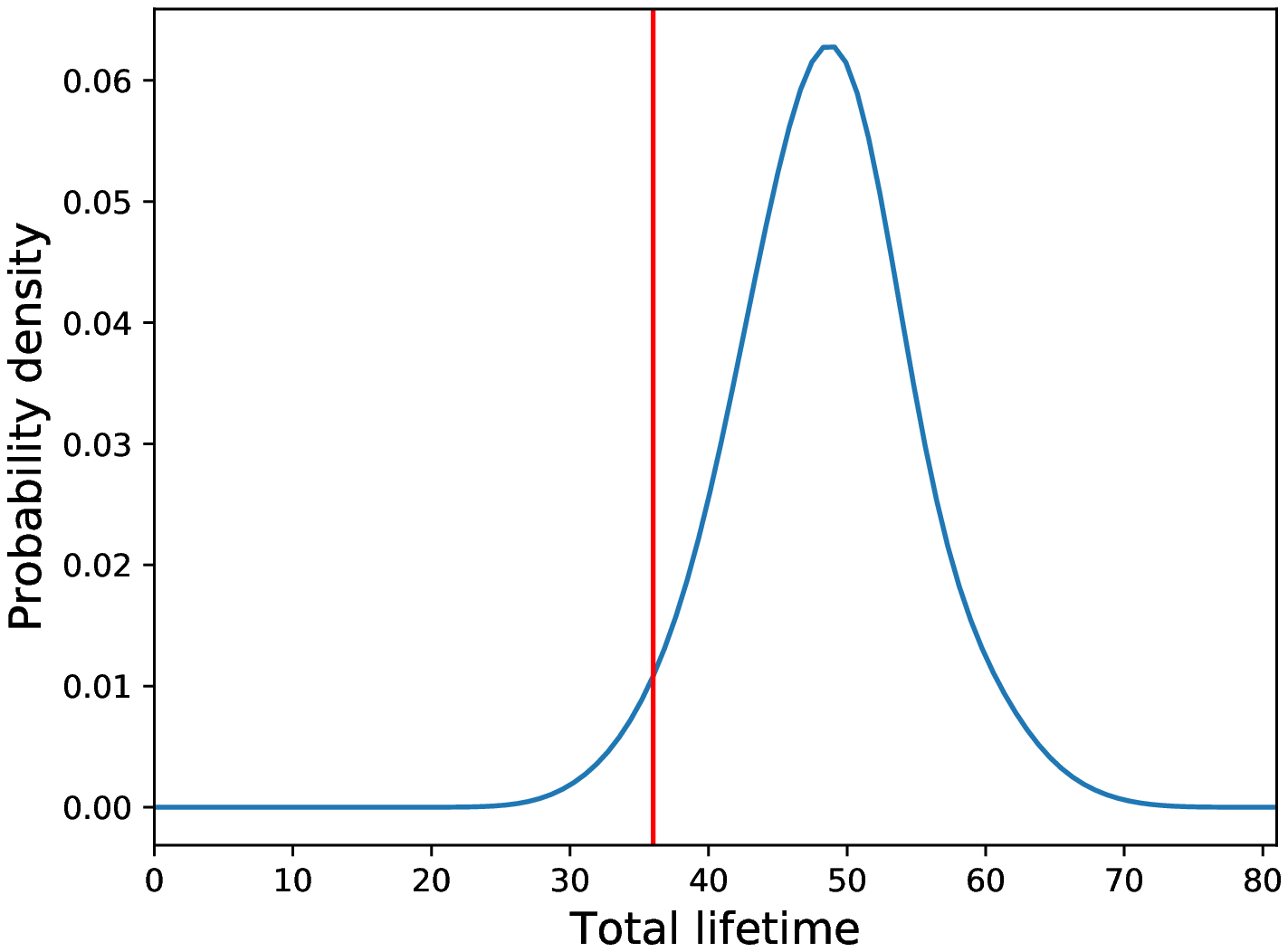}
    \caption{}
    \end{subfigure}
    \begin{subfigure}[b]{0.19\textwidth}
    \centering
    \includegraphics[width=\textwidth]{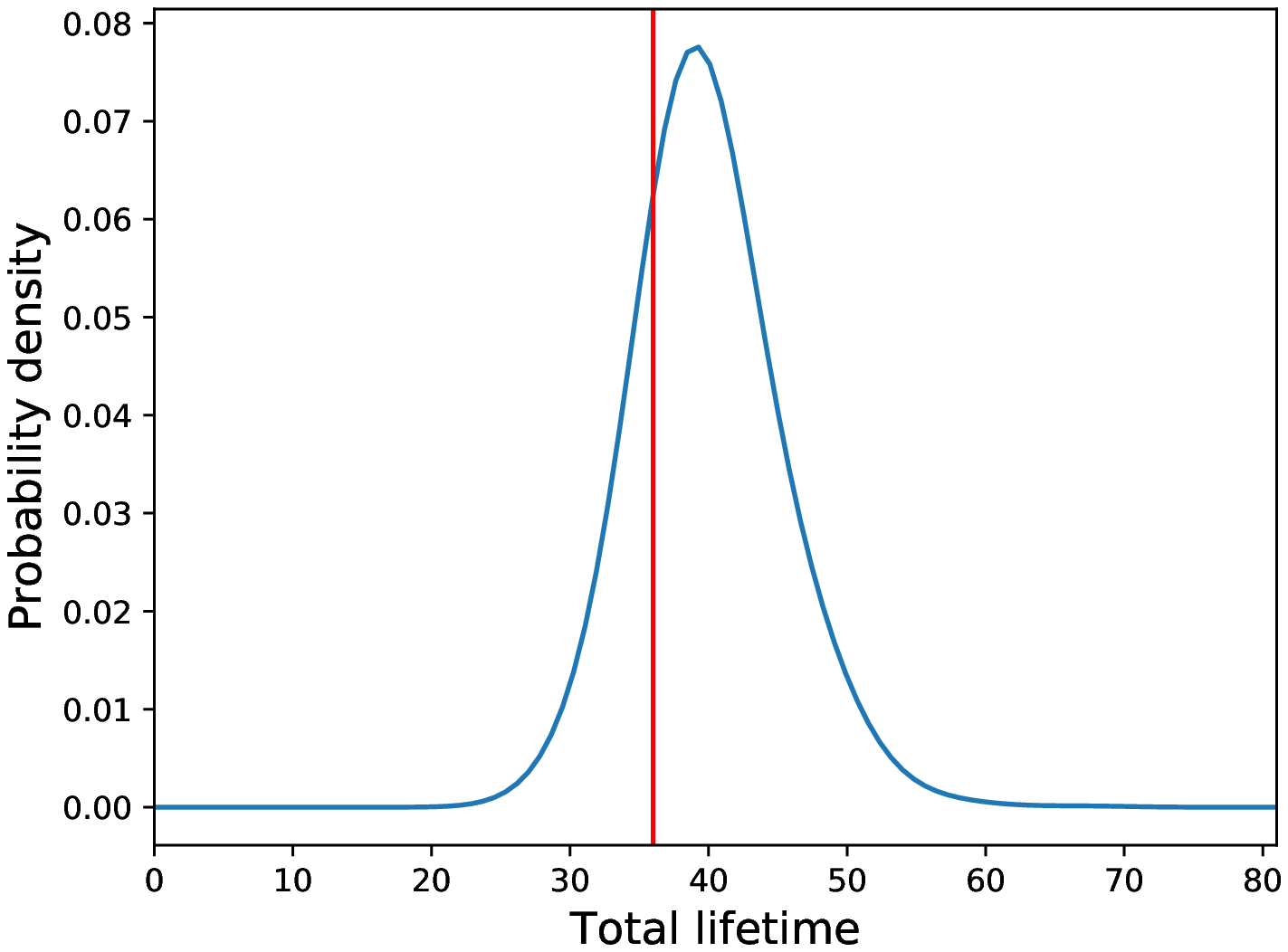}
    \caption{}
    \end{subfigure}
    \begin{subfigure}[b]{0.19\textwidth}
    \centering
    \includegraphics[width=\textwidth]{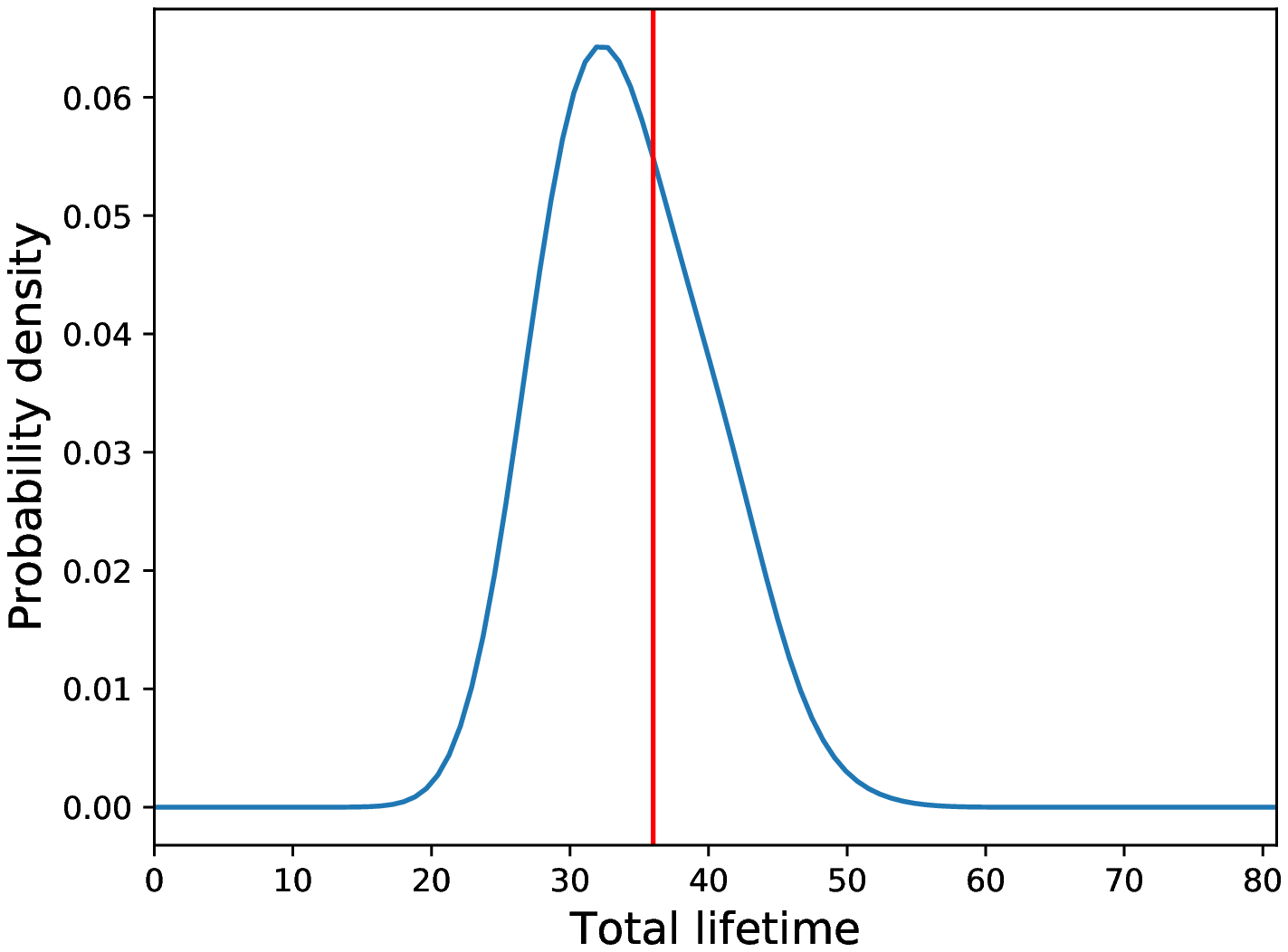}
    \caption{}
    \end{subfigure}
    \begin{subfigure}[b]{0.19\textwidth}
    \centering
    \includegraphics[width=\textwidth]{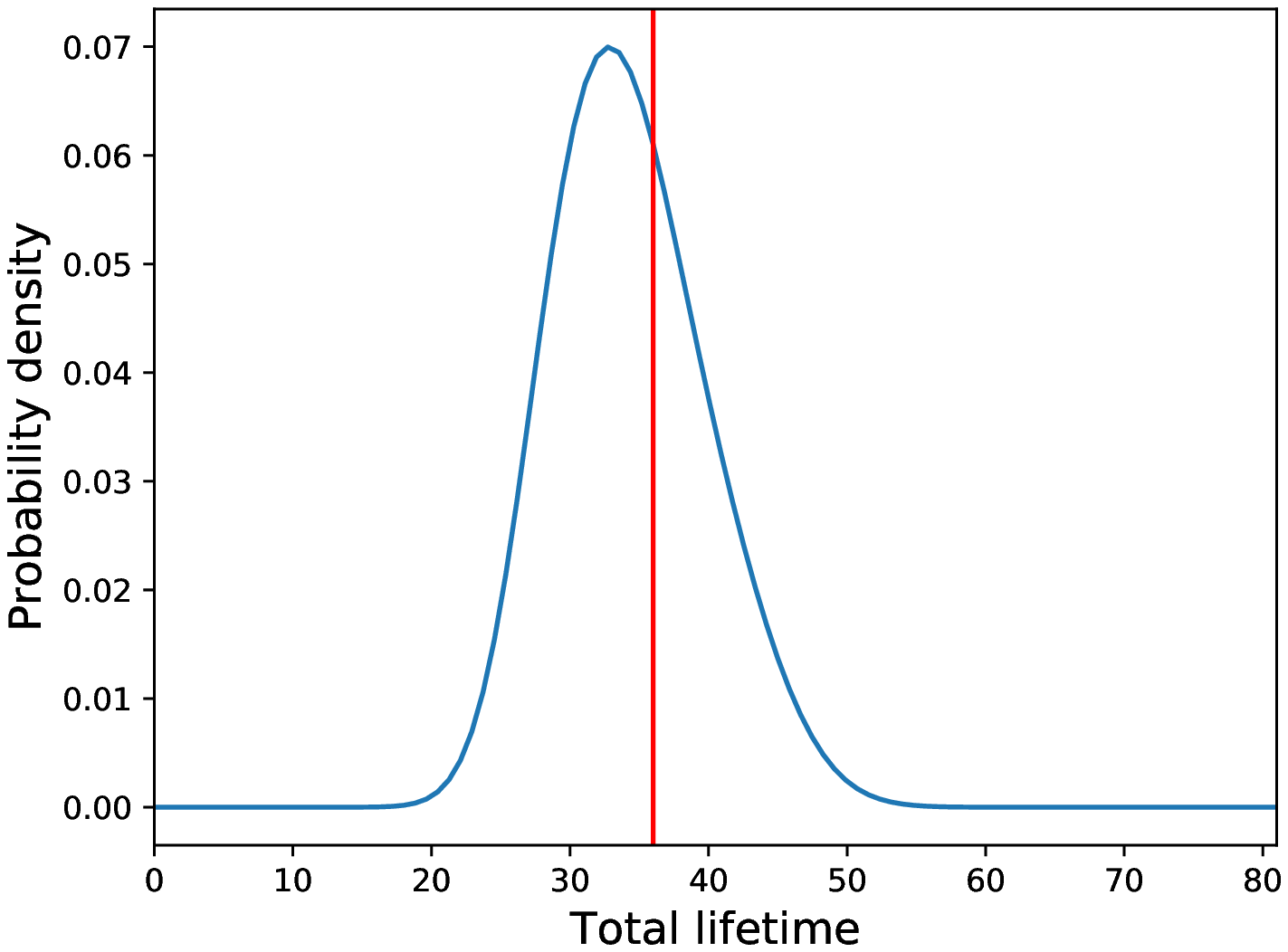}
    \caption{}
    \end{subfigure}
    \begin{subfigure}[b]{0.19\textwidth}
    \centering
    \includegraphics[width=\textwidth]{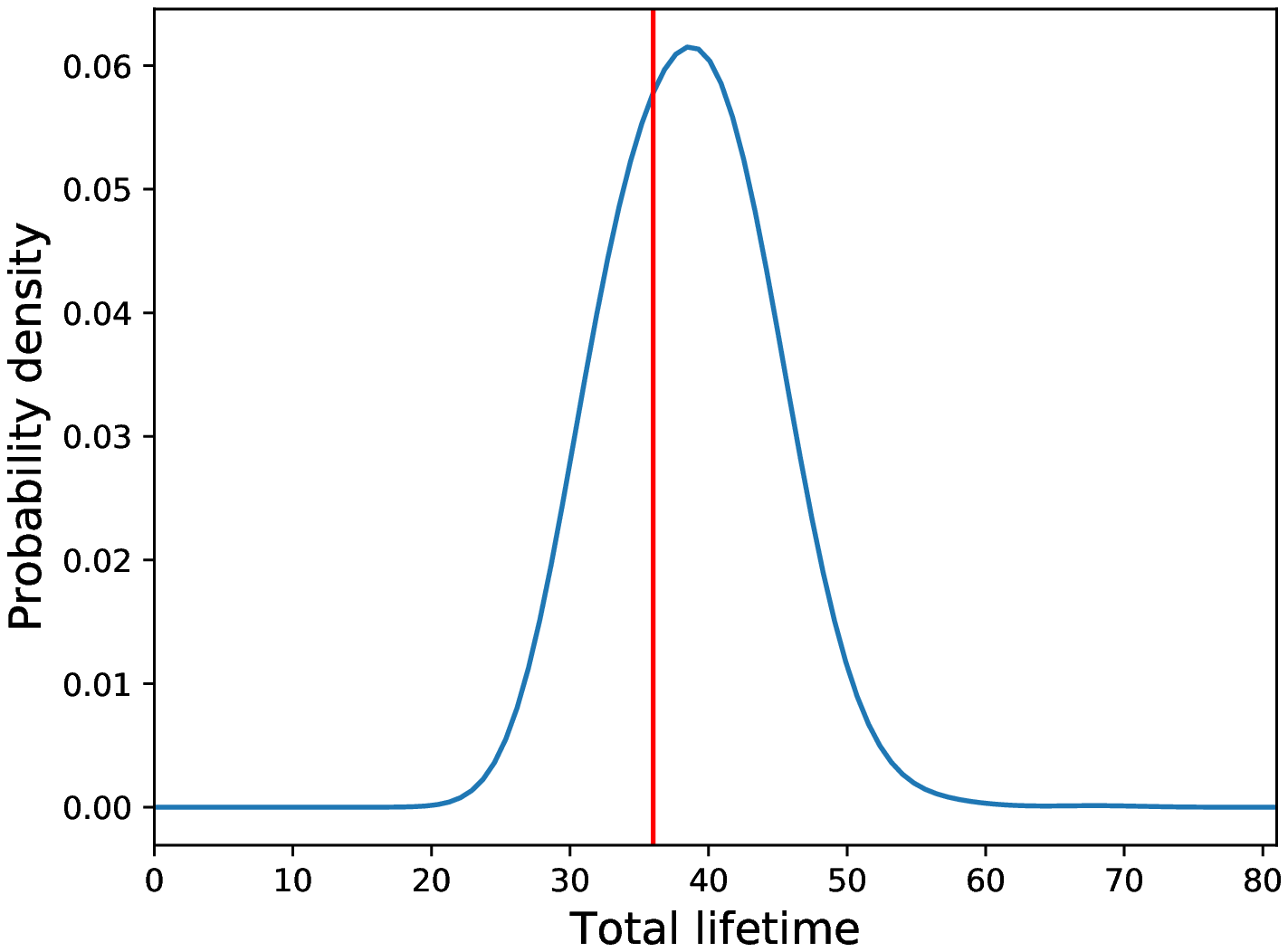}
    \caption{}
    \end{subfigure}
    \begin{subfigure}[b]{0.19\textwidth}
    \centering
    \includegraphics[width=\textwidth]{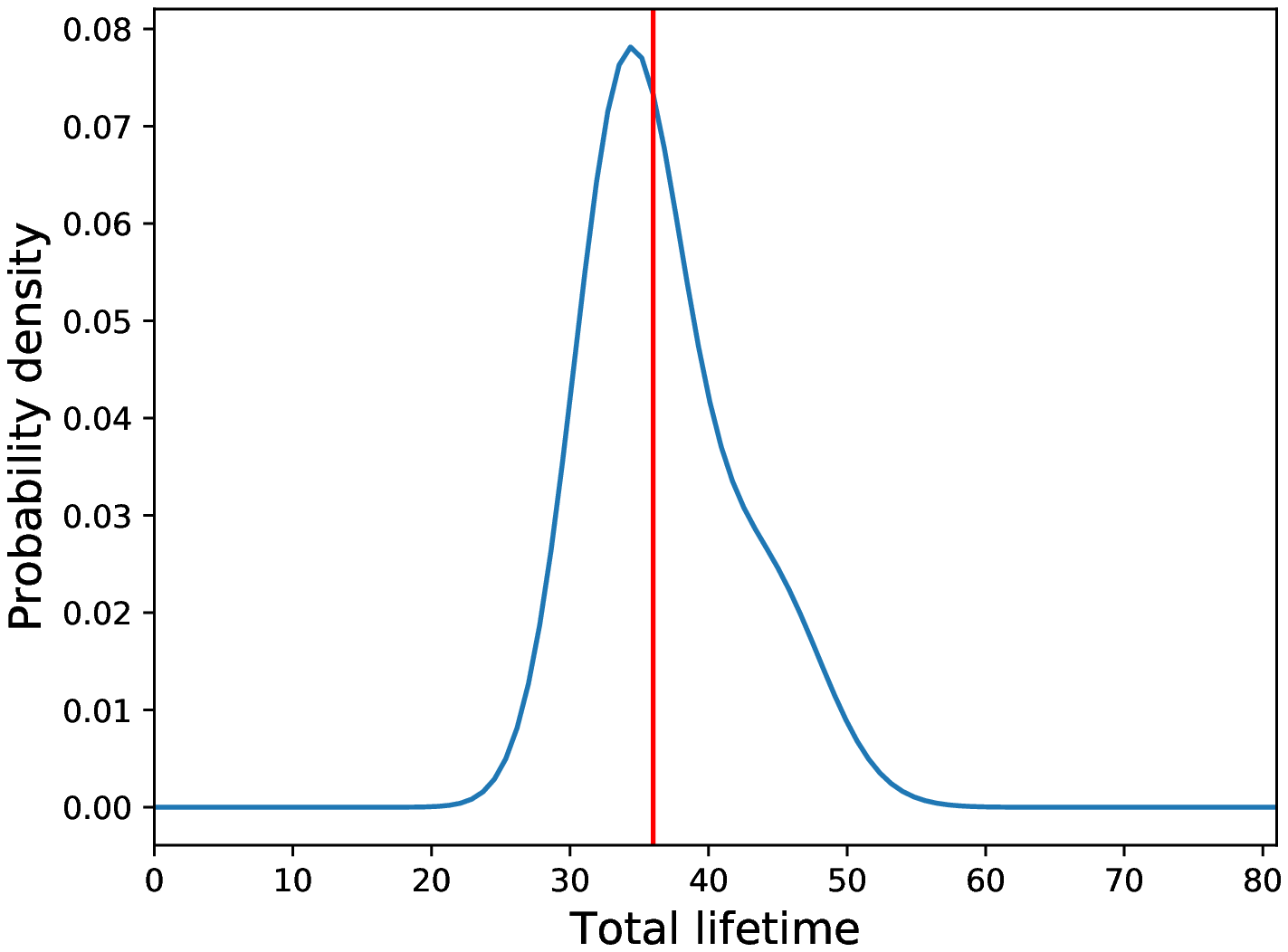}
    \caption{}
    \end{subfigure}
    \begin{subfigure}[b]{0.19\textwidth}
    \centering
    \includegraphics[width=\textwidth]{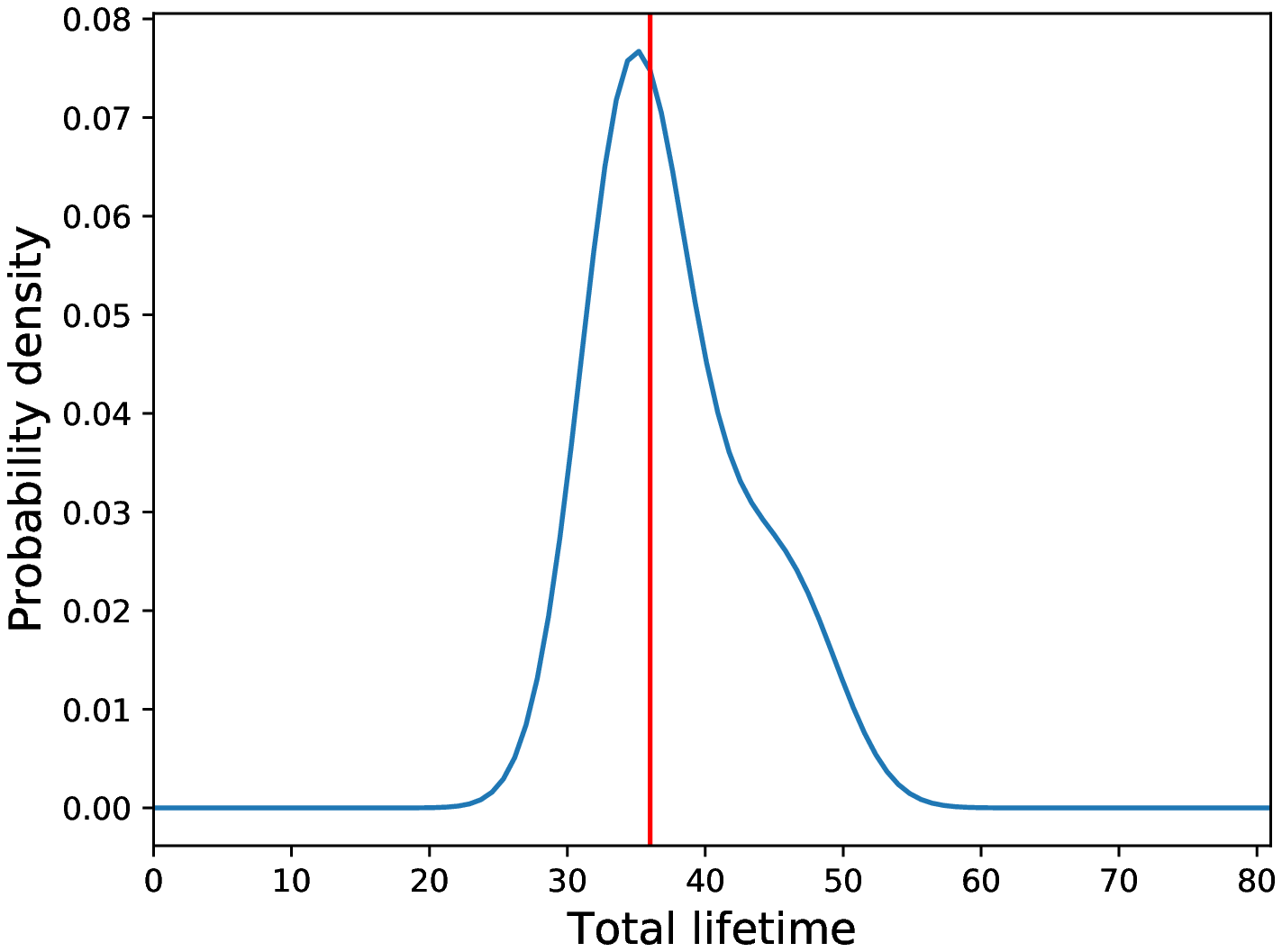}
    \caption{}
    \end{subfigure}
    \begin{subfigure}[b]{0.19\textwidth}
    \centering
    \includegraphics[width=\textwidth]{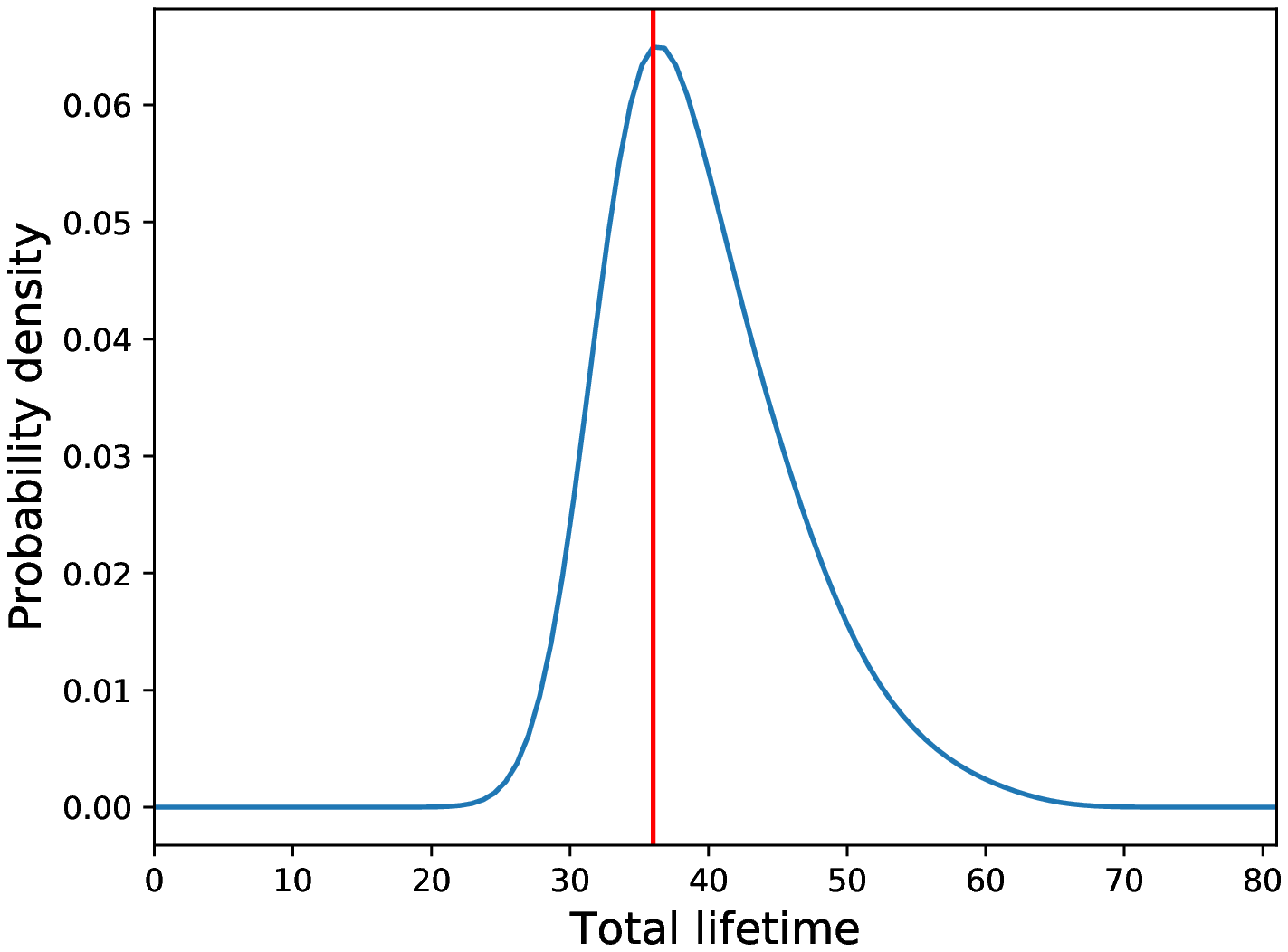}
    \caption{}
    \end{subfigure}
    \begin{subfigure}[b]{0.19\textwidth}
    \centering
    \includegraphics[width=\textwidth]{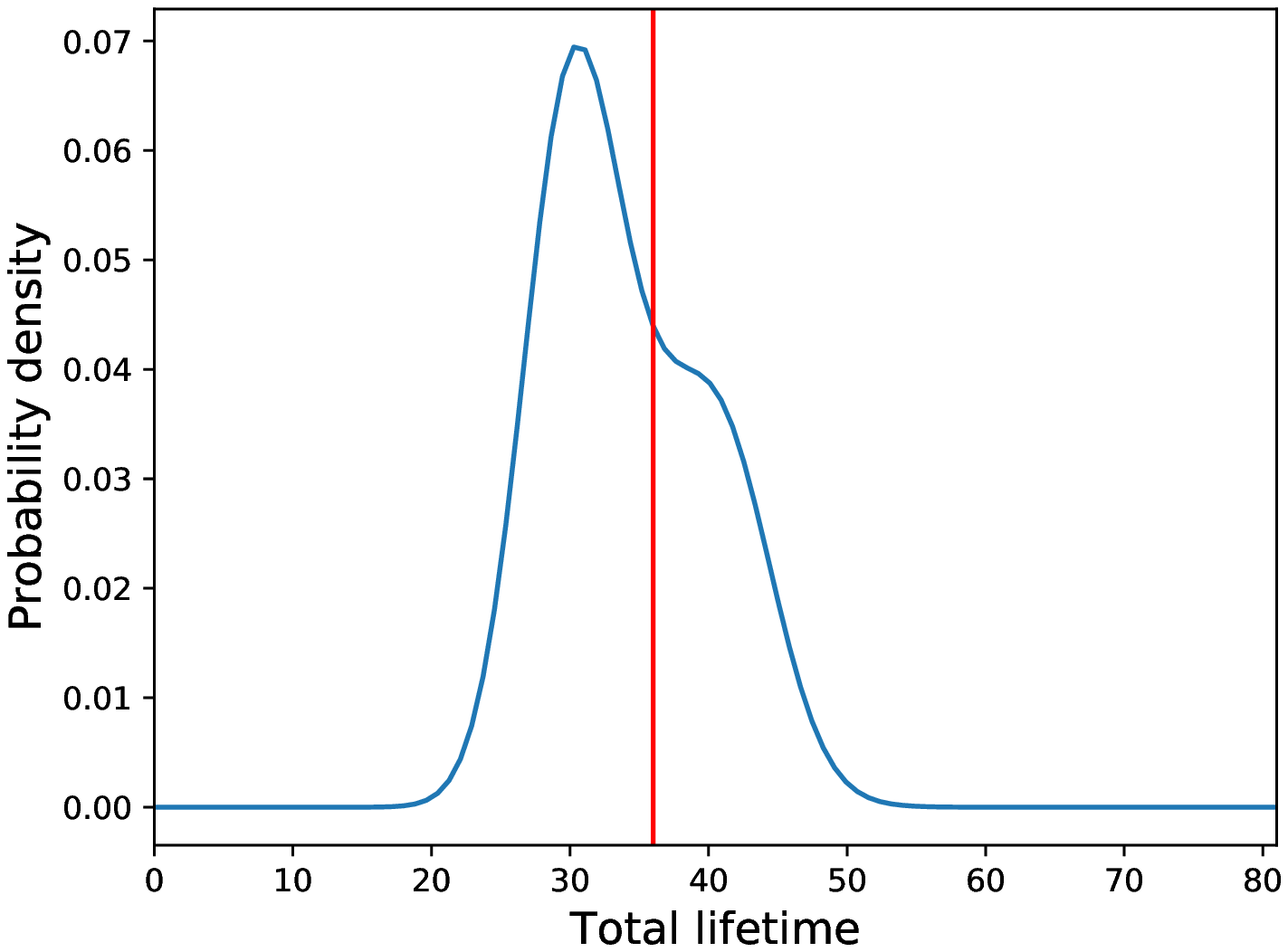}
    \caption{}
    \end{subfigure}
    \begin{subfigure}[b]{0.19\textwidth}
    \centering
    \includegraphics[width=\textwidth]{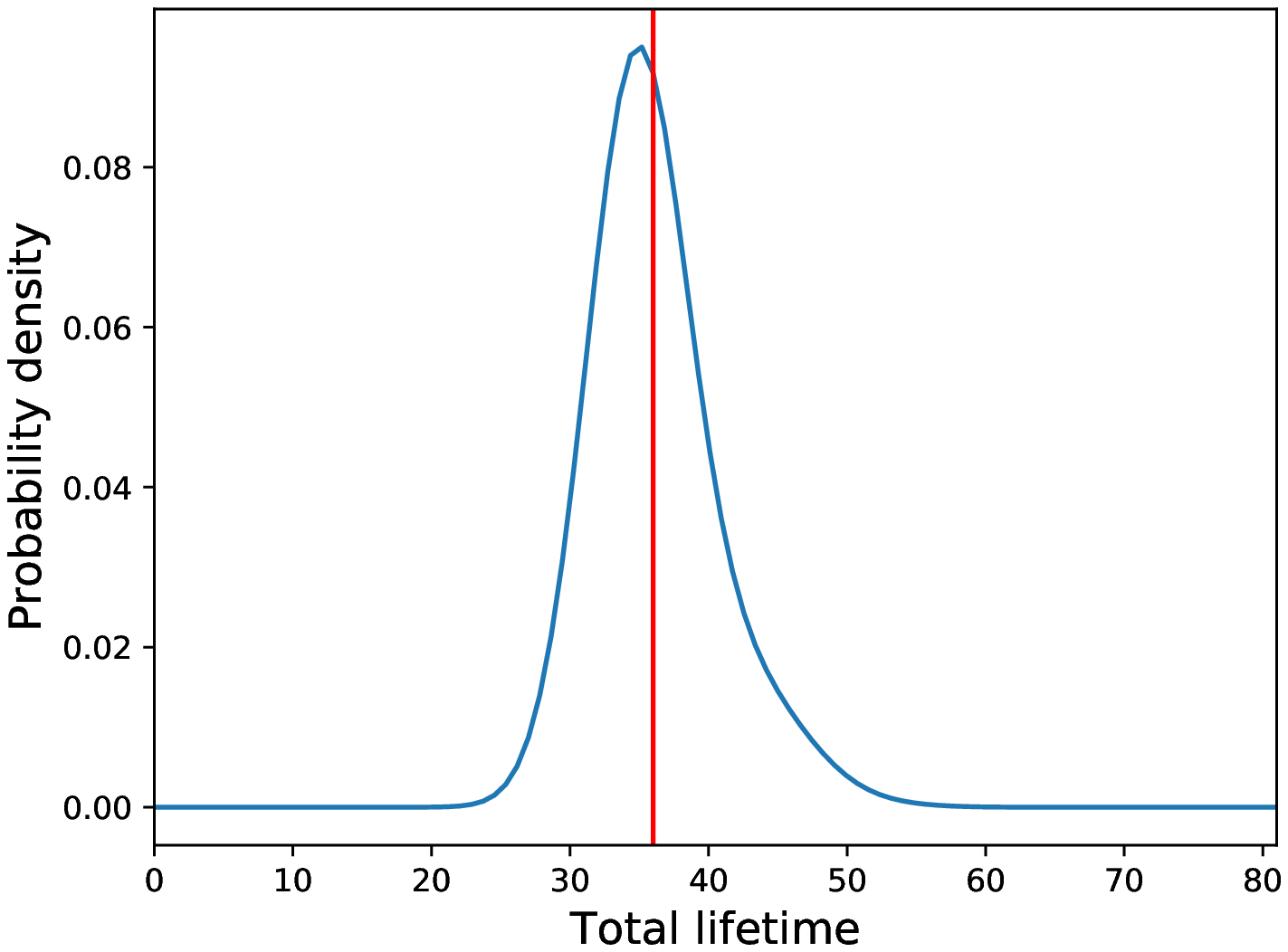}
    \caption{}
    \end{subfigure}
    \begin{subfigure}[b]{0.19\textwidth}
    \centering
    \includegraphics[width=\textwidth]{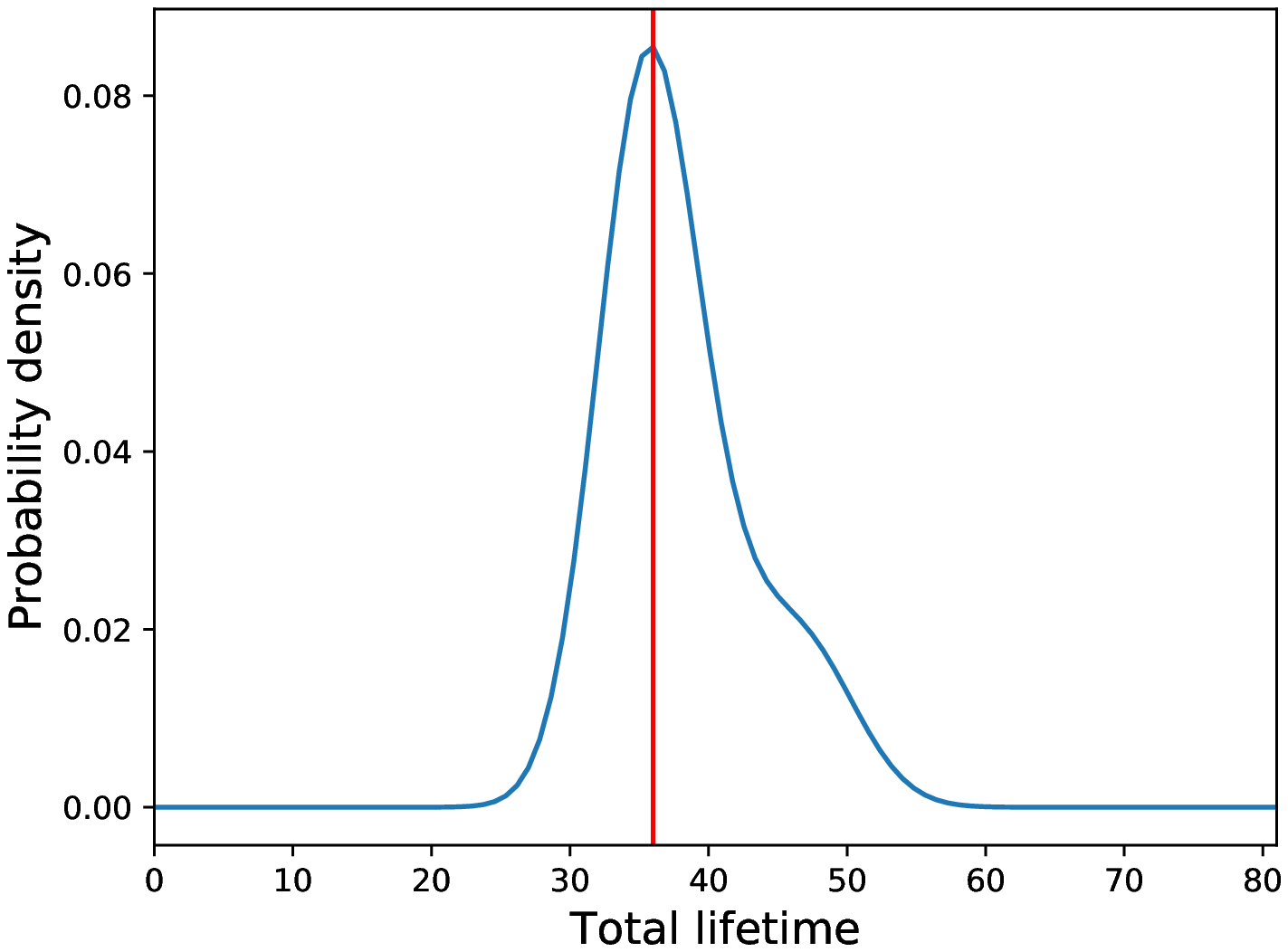}
    \caption{}
    \end{subfigure}
    \begin{subfigure}[b]{0.19\textwidth}
    \centering
    \includegraphics[width=\textwidth]{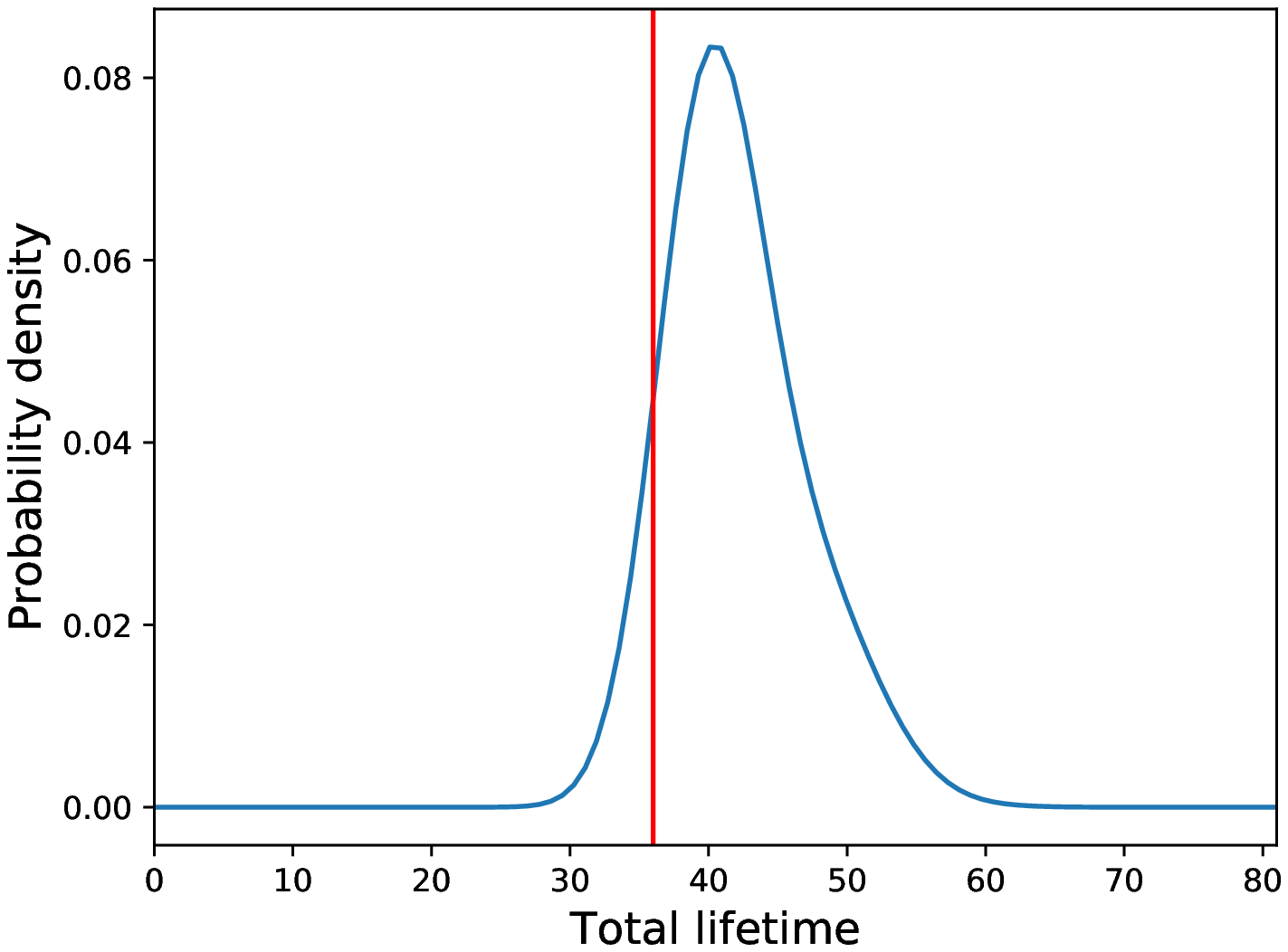}
    \caption{}
    \end{subfigure}
    \begin{subfigure}[b]{0.19\textwidth}
    \centering
    \includegraphics[width=\textwidth]{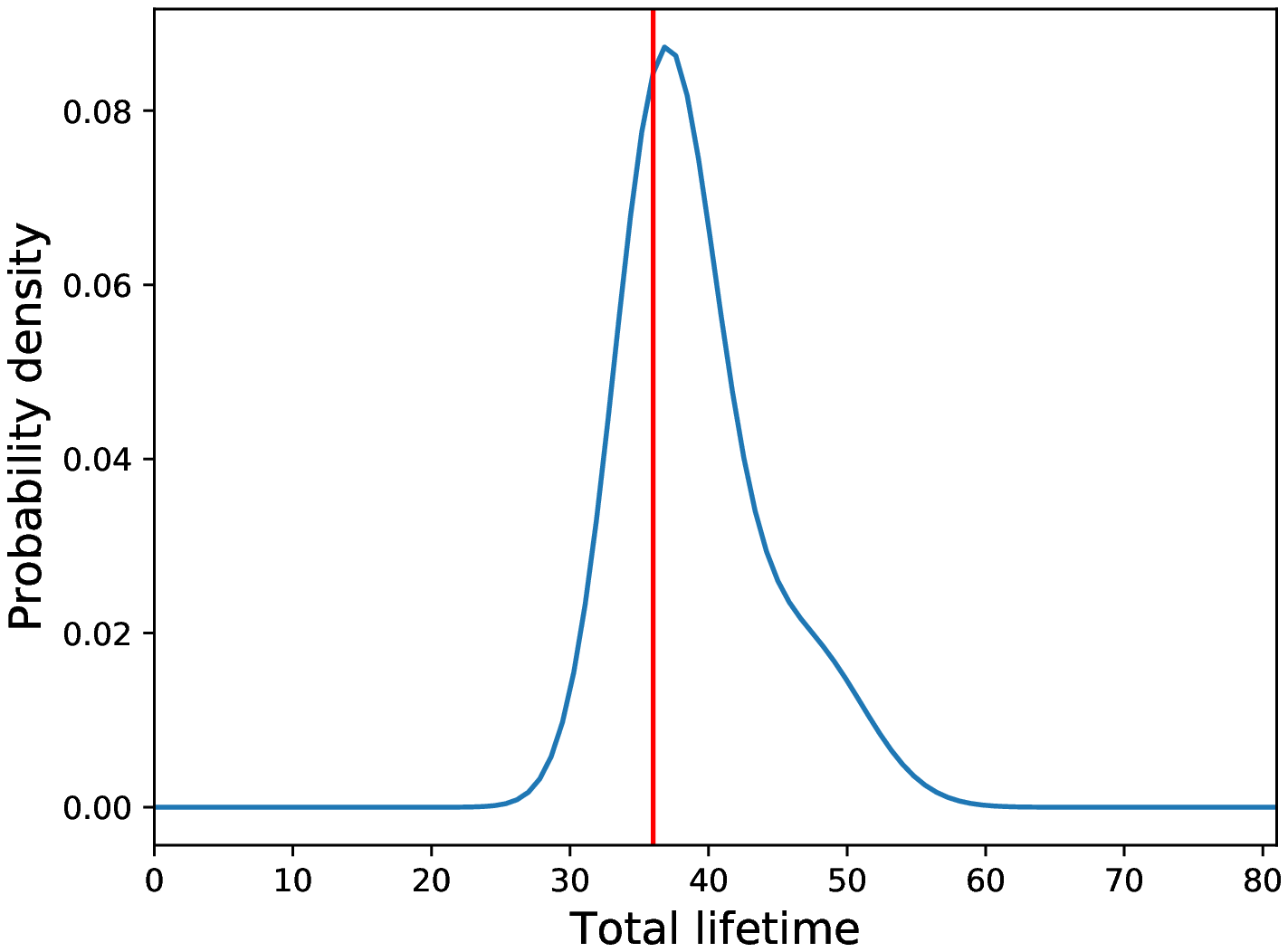}
    \caption{}
    \end{subfigure}
    \begin{subfigure}[b]{0.19\textwidth}
    \centering
    \includegraphics[width=\textwidth]{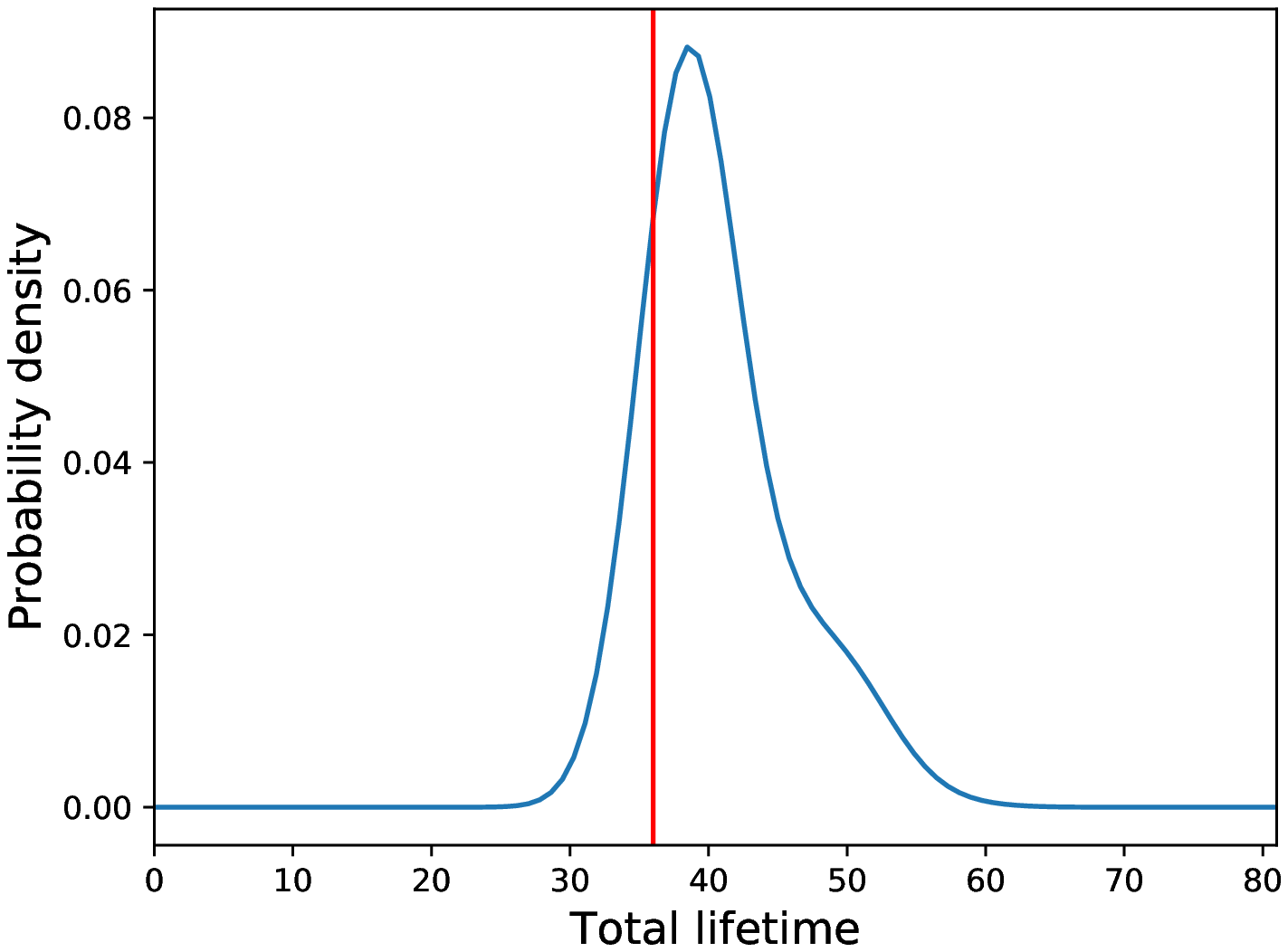}
    \caption{}
    \end{subfigure}
    \begin{subfigure}[b]{0.19\textwidth}
    \centering
    \includegraphics[width=\textwidth]{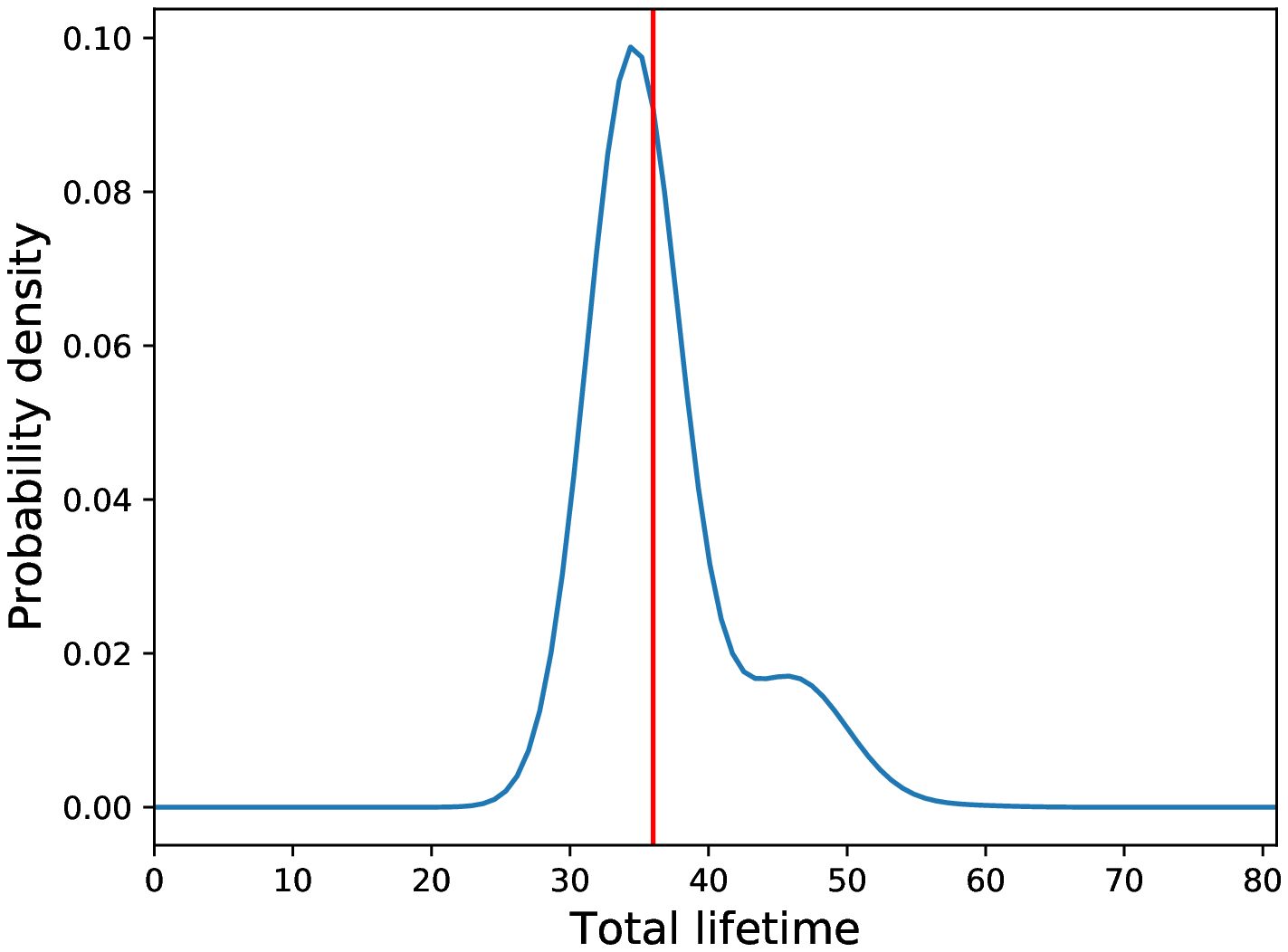}
    \caption{}
    \end{subfigure}
    \caption{\centering Real lifetime of tested structure (red vertical line) and evolution of predicted total life-time PDF using data from one more time-step in every figure; from left to right and from top to bottom.}
    \label{fig:pdf_evolution}
\end{figure}

After training the model, the algorithm which was described in the previous section was applied. In order to evaluate the ability of the algorithm to provide estimations of the remaining useful life of the structures, a second testing population of structures is considered using the same characteristics as the population used for training. The testing population was generated using the same characteristics for the random distributions of the structural parameters, as well as the same procedure of random damage evolution. Afterwards, for every timestep of every structure before its failure, the algorithm was used to estimate the remaining life of the structures. In Figure \ref{fig:pdf_evolution}, the evolution of the predicted PDF of the total lifetime of some structure is shown, as the algorithms acquires more information at every timestep. In the same Figure, the red vertical line indicates the real total lifetime of the specific structure.

In order to holistically evaluate the algorithm, the mean probability assigned to the real remaining lifetime at every timestep is calculated from,
\begin{equation}
    \label{eq:timegan_criterion}
    \pazocal{P} = \frac{1}{n_{test}}\sum_{i=1}^{n_{test}} p(T_{i})
\end{equation}
where $n_{test}$ is the total number of testing timesteps and $p(T_{i})$ is the real lifetime of the testing structure $i$. This metric for the testing population was equal to $0.058$. Following a naive approach and considering the PDF shown in blue in Figure \ref{fig:synth_real_comp} as a form \cite{PBSHMMSSP1} in order to assign a probability density to the remaining life of structures within the population, the metric of equation (\ref{eq:timegan_criterion}) is equal to $0.037$. This means that the TimeGAN algorithm proposed provides predictions with higher confidence. 
\newline

\section{Discussion}
\label{sec:discussion}

In the current work, a method to estimate the remaining useful life of structures within a PBSHM framework was presented. The algorithm is based on creating a Turing mirror model that is also a form of a homogeneous population. The model is about the evolution of damage that is observed on the structures. It is expected to learn the mechanism of the damage evolution and incorporate the uncertainty of the procedure. The core model used (TimeGAN) is a stochastic model that learns to generate artificial time-series according to rules learnt from a dataset incorporating any uncertainties that might be present in the procedure. The major advantage of the algorithm is that it learns to generate new points in the time-series taking into account previous states without a Markov assumption.

In the simulated example presented, the algorithm yielded quite encouraging results. Even though the way the data were generated may not resemble some realistic situation, similar damage evolution mechanisms are expected to be observed in fatigue damage situations. The algorithm is expected to learn to model the uncertain quantities of the procedure, such as environmental conditions and random events. Moreover, the algorithm is able to provide the user with potential future timeseries. Therefore, one has the chance to study them according to his knowledge and make appropriate decisions about the future of the structure.

% References
\bibliographystyle{unsrt}
\bibliography{On_paths_of_damage}

\end{document}